\documentclass[10pt,twocolumn,letterpaper]{article}

\usepackage{cvpr}              
\usepackage{multirow}
\usepackage{diagbox}
\usepackage[accsupp]{axessibility}  
\usepackage{mathtools}
\definecolor{cvprblue}{rgb}{0.21,0.49,0.74}
\usepackage[pagebackref,breaklinks,colorlinks,allcolors=cvprblue]{hyperref}

\def\paperID{} 
\def\confName{CVPR}
\def\confYear{2026}

\title{Towards Training-Free Scene Text Editing}

\author{Yubo Li$^{2,3,4}\thanks{Equal contribution. ~~\textsuperscript{\dag}Corresponding author.}$ , Xugong Qin$^{1,*}$, Peng Zhang$^{1,\dagger}$, Hailun Lin$^{2,3,}$,  Gangyan Zeng$^1$, Kexin Zhang$^1$\\
$^1$School of Cyber Science and Engineering, Nanjing University of Science and Technology\\
$^2$Institute of Information Engineering, Chinese Academy of Sciences\\
$^3$State Key Laboratory of Cyberspace Security Defense\\
$^4$School of Cyber Security, University of Chinese Academy of Sciences\\
{\tt\small liyubo2023@iie.ac.cn, qinxugong@njust.edu.cn}
}

\begin{document}
\maketitle
\begin{abstract}
Scene text editing seeks to modify textual content in natural images while maintaining visual realism and semantic consistency. Existing methods often require task-specific training or paired data, limiting their scalability and adaptability. In this paper, we propose TextFlow, a training-free scene text editing framework that integrates the strengths of Attention Boost (AttnBoost) and Flow Manifold Steering (FMS) to enable flexible, high-fidelity text manipulation without additional training. Specifically, FMS preserves the structural and style consistency by modeling the visual flow of characters and background regions, while AttnBoost enhances the rendering of textual content through attention-based guidance. By jointly leveraging these complementary modules, our approach performs end-to-end text editing through semantic alignment and spatial refinement in a plug-and-play manner. Extensive experiments demonstrate that our framework achieves visual quality and text accuracy comparable to or superior to those of training-based counterparts, generalizing well across diverse scenes and languages. This study advances scene text editing toward a more efficient, generalizable, and training-free paradigm. Code is available at \href{https://github.com/lyb18758/TextFlow}{https://github.com/lyb18758/TextFlow}
\end{abstract}    
\section{Introduction}
\label{sec:intro}

Scene Text Editing (STE) \cite{wu2019editing,roy2020stefann,qu2023exploring} aims to modify or replace text in natural images while preserving background and key visual attributes of the original text, including font style, color, size, and geometric layout. This task has broad practical value in applications such as image translation~\cite{wang2022pretrainingneedimagetoimagetranslation}, advertisement design~\cite{Zhao_2025}, content-aware image editing~\cite{yu2025skyreelstextfinegrainedfontcontrollabletext}, data augmentation for text  recognition~\cite{du2025mdiff4strmaskdiffusionmodel,qiao2021gaussian}, and other text-centric vision tasks~\cite {qin2019curved,qin2021fc2rn,qin2021mask,qin2023towards,tong2024granularity,qin2025cayn,zeng2024fdp,guo2025nldir,qin2025towards,guo2022units,guo2021waw}.

Generative models have evolved significantly, from early Generative Adversarial Networks (GANs) \cite{saxena2021generative,krishnan2023textstylebrush,yang2020swaptext,das2025faster,zhou2024explicitly} that faced training instability, to UNet-based diffusion models \cite{ho2020denoising,ji2023improving,chen2023diffute,tuo2023anytext,tuo2024anytext2,wang2023letter} that improved output fidelity and diversity, and further to Diffusion Transformers (DiT) \cite{esser2024scaling,flux2024,xie2025textflux,labs2025flux1kontextflowmatching,lan2025flux} that enhanced global semantic modeling through Multimodal Attention. These advances have propelled progress in STE, with methods such as DiffSTE \cite{ji2023improving}, AnyText~\cite{tuo2023anytext}, and textFlux \cite{xie2025textflux} demonstrating strong text-rendering performance.

\begin{figure}[t]
    \centering
    \includegraphics[width=1\linewidth]{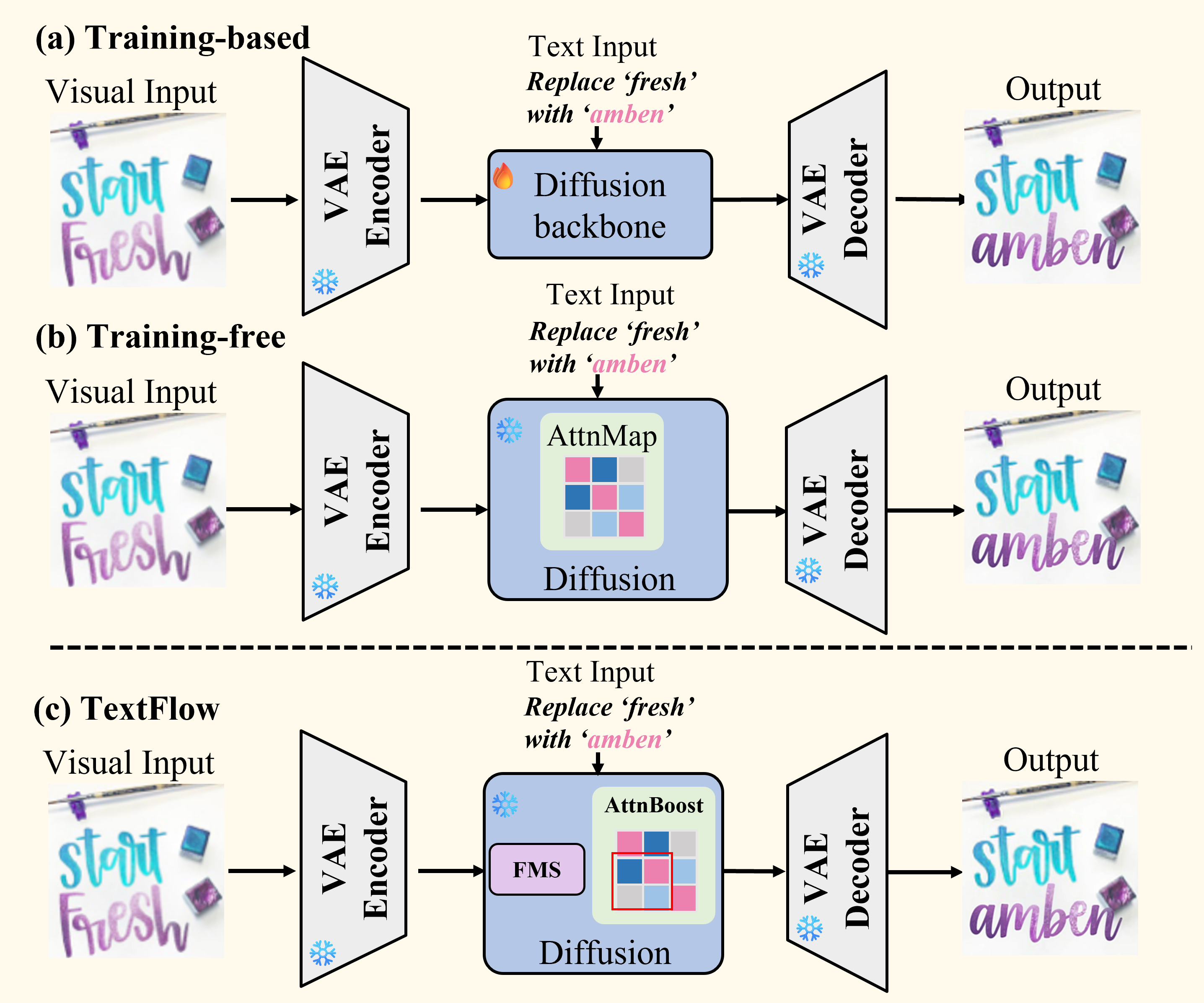}
    \vspace{-20px}
    \caption{Comparison of the pipelines between training-based and training-free methods for scene text editing. Training-based methods require large-scale, high-quality paired data that require high computing resources. The training-free method mostly focuses on the attention map for general objects, but ignores the text accuracy and style consistency.}
    \label{fig:1}
    \vspace{-20px}
\end{figure}

However, a fundamental trade-off exists between adaptability and editing quality. Training-based methods, like Fig.~\ref{fig:1}(a), require large-scale, high-quality paired data, which is scarce in practice. While synthetic data can supplement training, it often limits generalization to diverse real scenes. Additionally, these approaches demand substantial computational resources, restricting their practical use. Training-free methods, as shown in Fig.~\ref{fig:1}(b), leverage pre-trained models without fine-tuning, with many approaches utilizing attention manipulation for editing tasks. While effective for general object editing, these methods face particular challenges in scene text editing. Preserving precise typographic and structural details in complex scenes with diverse backgrounds, fonts, or layouts remains challenging for attention-based methods, often resulting in visual artifacts and character distortions.

A key limitation of training-free methods lies in their phase-dependent controllability, which arises from the non-uniform signal-to-noise ratio across diffusion timesteps.
During early denoising, existing techniques fail to preserve the structural and stylistic foundations, resulting in unstable editing trajectories. In later stages, inadequate semantic and spatial guidance leads to textual inaccuracies, such as character duplication, missing elements, or distortion, thereby hindering coherent text generation.


To address these challenges, we propose TextFlow, a training-free framework for scene text editing. As illustrated in Fig.~\ref{fig:1}(c), TextFlow introduces phase-aware guidance that separately optimizes style preservation and textual accuracy. Specifically, it operates in two phases: the first employs a Flow Manifold Steering (FMS) module to maintain style consistency, while the second leverages an Attention Boost (AttnBoost) mechanism to improve textual accuracy. Despite requiring no training, our method narrows the performance gap with training-based approaches, achieving competitive editing quality through a single forward pass without task-specific fine-tuning, paired datasets, or resource-intensive retraining. This makes TextFlow both efficient and practical for real-world applications.
The main contributions of this work can be summarized as follows:

\begin{itemize}
    \item We introduce \textbf{Flow Manifold Steering (FMS)} module, which operates source and target conditions in the latent space, guiding the denoising trajectory to maintain structural and stylistic consistency from the denoising steps.
    \item We propose an \textbf{Attention Boost (AttnBoost)} mechanism that leverages attention maps to enhance fine-grained text rendering. By dynamically amplifying text-relevant regions during sampling, AttnBoost significantly improves textual accuracy and semantic alignment.
    \item Through extensive experiments on benchmark datasets, we demonstrate that TextFlow achieves state-of-the-art performance in both visual quality and textual correctness, without any task-specific fine-tuning.
\end{itemize}

\begin{figure*}[t]
    \centering
    \includegraphics[width=0.9\linewidth]{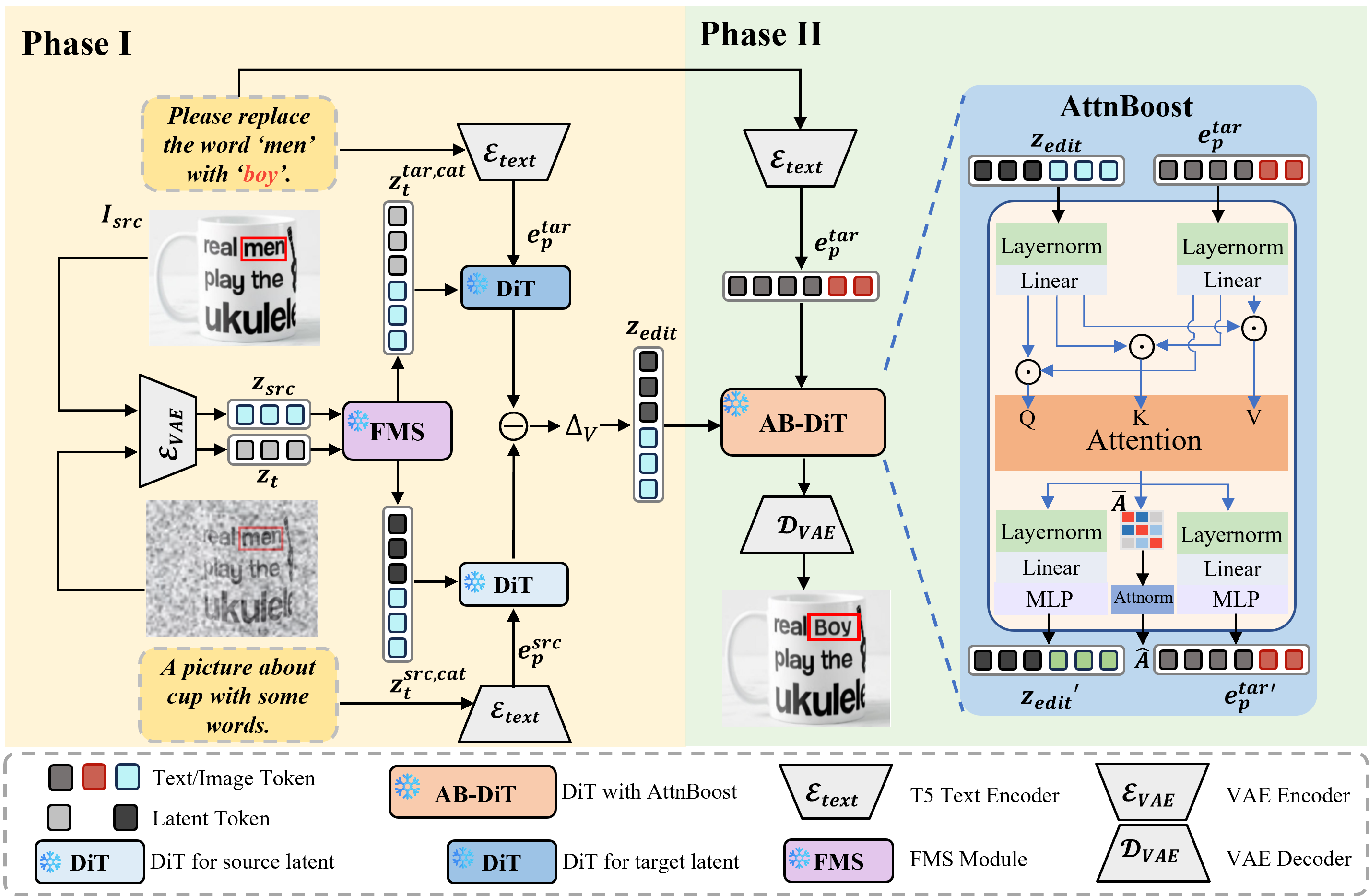}
    \caption{The overall framework of TextFlow. In the first phase, the source image is encoded into latent representations $\mathbf{z}_t$ and $\mathbf{z}_{src}$ via the VAE encoder, which are subsequently processed by the FMS module to generate concatenated representations $\mathbf{z}^{src,cat}_t$ and $\mathbf{z}^{tar,cat}_t$. These representations, along with their corresponding text embeddings $\mathbf{e}^{src}_p$ and $\mathbf{e}^{tar}_p$, are fed into parallel DiT blocks to compute the velocity field differential $\Delta_V$, ultimately producing the edited latent representation $\mathbf{z}_{edit}$; In the second phase, $\mathbf{z}_{edit}$ and the target embedding $\mathbf{e}^{tar}_{p}$ are processed by the AttnBoost DiT (AB-DiT), where concatenation and self-attention operations generate refined text-to-image attention maps that enhance textual rendering accuracy through spatial-aware amplification.}
    \label{fig:2}
\end{figure*}

\section{Related Work}
\label{sec:relat}

\subsection{Diffusion-Based Scene Text Editing}

The widespread application of the UNet-based diffusion model in image editing has driven the development of STE. 

DiffSTE~\cite{ji2023improving} employs a dual-encoder design with character and instruction encoding to learn the mapping from textual instructions to corresponding images with specified styles in the background; TextDiffuser~\cite{chen2023textdiffuserdiffusionmodelstext} systematically decouples layout planning from content generation by employing a dual-stage framework; DiffUTE~\cite{chen2023diffute} utilizes character glyphs and text positions from the source image as auxiliary information to provide better control during character generation; UDiffText~\cite{zhao2024udifftext} leverages large-scale training data and text embeddings to improve text-based image editing; AnyText~\cite{tuo2023anytext} encodes auxiliary information such as text glyphs, positions, and mask images into a latent space to assist in text generation and editing; AnyText2~\cite{tuo2024anytext2} proposes a WriteNet+AttnX architecture, enabling the model to focus more on font and color attributes; DreamText~\cite{wang2025dreamtext} effectively mitigates issues of character repetition, omission, and distortion encountered by existing methods; TextCtrl~\cite{zeng2024textctrl} decomposes the prerequisites of STE into fine-grained style disentanglement and glyph structure representation, integrating style-structure guidance with diffusion models to enhance rendering accuracy and style fidelity; GlyphMastero~\cite{wang2025glyphmasteroglyphencoderhighfidelity} targets editing tasks with complex characters, such as Chinese, by combining local character-level features and global text-line structures.

To further enhance generation performance, recent studies integrate large-scale transformer architectures as the backbone of diffusion models, resulting in advanced models like DiT~\cite{peebles2023scalable}. Stable Diffusion 3~\cite{esser2024scaling} and FLUX~\cite{flux2024}, both based on the flow matching method, have extended the DiT architecture to MM-DiT to achieve superior generation quality. Their subsequent open-source release has provided a significantly more robust foundation for STE. textFlux~\cite{xie2025textflux} eliminates the need for OCR encoders; FLUX-Text~\cite{lan2025flux} enhances glyph understanding and generation through lightweight Visual and Text Embedding Modules; Flux-kontext~\cite{labs2025flux1kontextflowmatching} generates novel output views by incorporating semantic context from text and image inputs; Qwen-image~\cite{wu2025qwen} separately feed the original image into Qwen2.5-VL and the VAE encoder to obtain semantic and reconstructive representations; HunYuanImage3.0~\cite{cao2025hunyuanimage} unifies multimodal understanding and generation within an autoregressive framework. Moreover, GPT-4o Image~\cite{openai2024gpt4ocard}, Gemini 2.5 Flash Image, and Blip3o-NEXT~\cite{chen2025blip3onextfrontiernativeimage} leverage a hybrid Diffusion-Autoregressive architecture to attain state-of-the-art capabilities in image understanding, generation, and editing.

While obtaining exceptional performance on STE tasks, existing methods typically demand considerable resources to solve the challenging problem of editing.

\subsection{Training-Free Image Editing}

Benefiting from the rapid advancement of the DiT backbone and flow matching techniques, foundation models have demonstrated significantly enhanced generation and editing capabilities alongside robust general-purpose performance. Building upon this progress, there is increasing research interest in exploring training-free methods to further improve the image editing proficiency of these models. 

Stable Flow~\cite{Avrahami_2025_CVPR} introduce an improved image inversion method for flow models to enable image editing; CannyEdit~\cite{xie2025canny} propose selective canny control and dual-prompt guidance to balance text adherence in edited regions, context fidelity in unedited areas, and seamless integration of edits; ICEdit~\cite{zhang2025context} adopt a diptych framework for both T2I-DiT and inpainting-DiT to achieve in-context editing; KV-Edit~\cite{zhu2025kv} uses KV cache in DiTs to maintain background consistency, ultimately generating new content that seamlessly integrates with the background within user-provided regions; RF-Solver~\cite{wang2024taming} proposes a novel training-free sampler that effectively enhances inversion precision by mitigating errors in the ordinary differential equation (ODE) solving process of rectified flow; FlowEdit~\cite{kulikov2024flowedit} constructs a direct path between the source and target distributions by breaking away from the editing-by-inversion paradigm; LanPaint~\cite{zheng2025lanpainttrainingfreediffusioninpainting} propose a training-free, asymptotically exact partial conditional sampling methods for ODE-based and rectified flow models.

Furthermore, building upon these general frameworks, visual text rendering and generation have also seen significant advancements. Specifically, AMO~\cite{hu2025amo} introduce an overshooting sampler for pretrained rectified flow (RF) models, by alternating between over-simulating the learned ODE and reintroducing noise, which improves the text rendering accuracy without compromising image quality; TextCrafter~\cite{du2025textcrafter} focusing on complex visual text generation, employs a progressive strategy to decompose complex visual text into distinct components while ensuring robust alignment between textual content and its visual carrier.

These methods perform outstandingly in general editing and text rendering. However, for the STE task, there is a distinct lack of research dedicated to training-free methods.

\vspace{-5px}
\section{Methodology}
\vspace{-5px}
\label{sec:metho}
In this section, we explore training-free editing capabilities within DiT generative models and propose our fusion edit framework for scene text editing.
Our fusion framework is based on the flow matching architecture, a continuous-time generative model that aims to learn a velocity field 
\(v_t(x)\), such that the ODE trajectory defined by this field maps noise \(\epsilon\sim\mathcal{N}(0,I)\) to the data sample \(x\).
Building upon FLUX-Kontext \cite{labs2025flux1kontextflowmatching}  implemented via flow matching, our approach introduces an innovative two-phased strategy, achieving high-precision scene text editing with low computational cost.

\subsection{Overall Framework}
The overall pipeline of our proposed TextFlow for denosing steps is illustrated in Fig.~\ref{fig:2}. Our core insight is to decouple the complex STE task into two complementary phases, each governed by a specialized mechanism to address its unique challenges: \textbf{style preservation}  and \textbf{detail rendering} during the denoising step.

Given a source image \( I_{src} \) with its corresponding caption \( T_{src} \) and a target text prompt \( T_{tar} \), the process begins by encoding the image into a latent representation to \( \mathbf{z}_{t}\) and \(\mathbf{z}_{src}\),  processing both texts through a text encoder to obtain their embeddings \( \mathbf{e}^{src}_p \) and \( \mathbf{e}^{tar}_p \). The denoising trajectory, governed by a pre-trained flow matching model, is then strategically manipulated by our two novel components:

\begin{itemize}
    \item \textbf{FMS module:} Operating in the first phase, as shown in Fig.~\ref{fig:2}, this module is responsible for establishing and preserving the foundational style and structure of the source image. The outputs compute a velocity field differential \( \mathbf{V}_{\Delta} \) between the source and target trajectories in the latent space and apply a controlled shift, ensuring that the global attributes (e.g., font style, background texture) are coherently retained early in the generation process.
    \item \textbf{AttnBoost mechanism:} Activated in the second phase, as shown in Fig.~\ref{fig:2}, this mechanism ensures the accurate spelling, legibility, and semantic alignment of the generated text. It extracts and processes the attention maps from the double-stream transformer block, generating a fine-grained guidance signal \( \hat{A} \) that directs the scheduler to render text details that precisely match the target description \( T_{tar} \).
\end{itemize}

\begin{figure}
    \centering
    \includegraphics[width=0.55\linewidth]{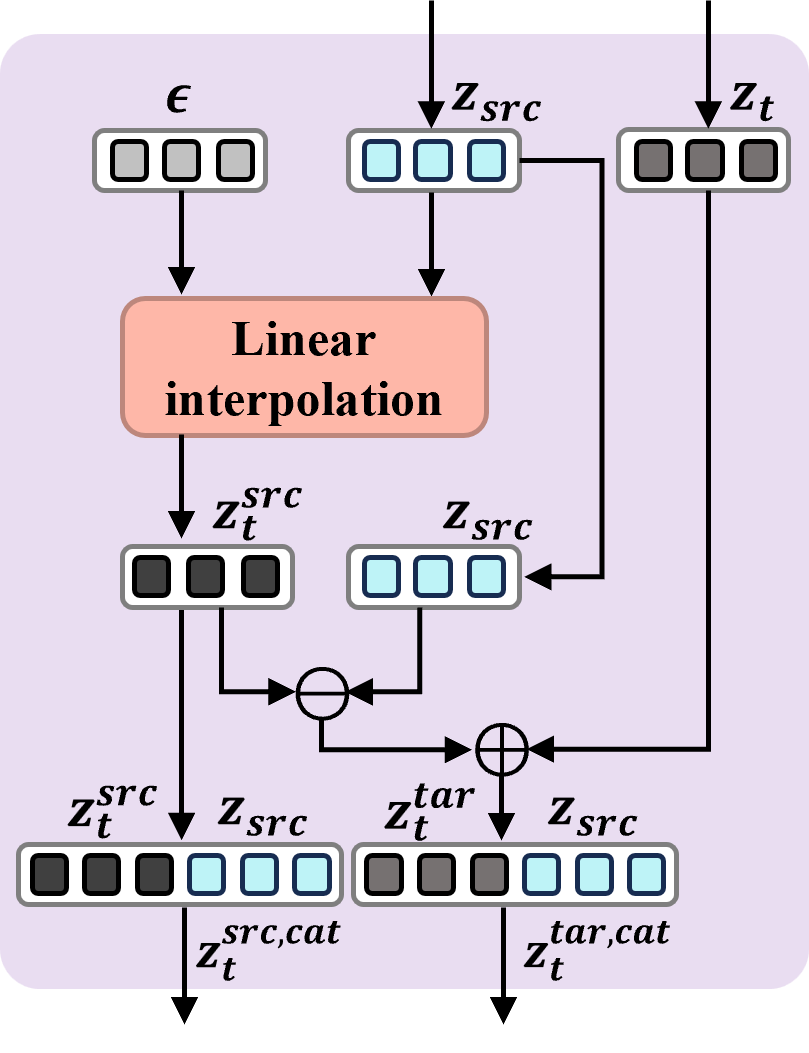}
    \vspace{-10px}
    \caption{Illustration of the proposed FMS Model. The latent representations $\mathbf{z}_t$ and $\mathbf{z}_{src}$ are processed with random noise $\epsilon$ through linear interpolation and vector arithmetic operations to maintain style consistency.}
    \vspace{-10px}
    \label{fig:3}
\end{figure}

\subsection{Style Preservation with FMS}

During the first phase of the denoising cycle, as shown in Fig.~\ref{fig:3}, we introduce the FMS module to achieve robust style preservation. This approach operates by manipulating trajectories in the latent space, ensuring structural integrity while accommodating stylistic transformations throughout the editing process.

The core framework of FMS consists of the following three steps. First, we define the parameter controlling noise injection intensity:
\begin{equation}
t_i = \sigma_{\text{step}}[i],
\end{equation}
where $t_i$ represents the noise level at the current timestep, and $\sigma_{\text{step}}$ denotes the standard deviation parameter from the diffusion scheduler.

Next, we construct the noise-injected source latent representation:
\begin{equation}
\mathbf{z}_t^{\text{src}} = (1 - t_i) \cdot \mathbf{z}_{\text{src}} + t_i \cdot \epsilon,
\end{equation}
where $\mathbf{z}_{\text{src}}$ is the original latent representation of the source image, $\mathbf{z}_t^{\text{src}}$ is the noise-injected latent state, and $\epsilon$ represents random noise following a standard normal distribution.

We then correct the target latent representation through differential geometric transformation:
\begin{equation}
\mathbf{z}_t^{\text{tar}} = \mathbf{z}_t + (\mathbf{z}_t^{\text{src}} - \mathbf{z}_{\text{src}}),
\end{equation}
where $\mathbf{z}_t$ is the current latent state of target generation, and $\mathbf{z}_t^{\text{tar}}$ is the corrected target representation. The differential term $(\mathbf{z}_t^{\text{src}} - \mathbf{z}_{\text{src}})$ precisely captures the geometric offset induced by noise injection.

To integrate information, we concatenate the processed states:
\begin{equation}
    \mathbf{z}_t^{\text{src,cat}} = \text{Concat}(\mathbf{z}_t^{\text{src}}, \mathbf{z}_t),
\end{equation}
\begin{equation}
    \mathbf{z}_t^{\text{tar,cat}} = \text{Concat}(\mathbf{z}_t^{\text{tar}}, \mathbf{z}_t).
\end{equation}

Furthermore, we compute the trajectory-shifting vector field for fine-grained control:
\begin{equation}
\mathbf{V}_{\Delta} = \mathcal{F}\left(\mathbf{z}_t^{\text{src,cat}},\mathbf{z}_t^{\text{tar,cat}}, \mathbf{e}^{\text{src}}_p, \mathbf{e}^{\text{tar}}_p\right),
\end{equation}
\begin{equation}
    \mathcal{F}\coloneqq\Phi(z_t^{tar,cat}, e^{tar}_p) - \Phi(z_t^{src,cat},e^{src}_p),
\end{equation}
where $\mathcal{F}$ is the velocity field computation function that performs cross-modal feature alignment between source and target embeddings. $\Phi$ represents the standard DiT backbone. Based on this differential, we apply trajectory shifting as follows:
\begin{equation}
\mathbf{z}_{\text{edit}} = \mathbf{z}_t + \mathbf{V}_{\Delta} \cdot \left(t_{i-1} - t_i\right),
\end{equation}
where $t_{i-1}$ and $t_i$ represent adjacent noise levels in the diffusion process.

This mathematical framework embeds structural preservation constraints into the generation trajectory through rigorous geometric operations, ensuring style coherence while supporting flexible text adaptation, thereby providing a theoretical foundation for training-free scene text editing.

\subsection{Detail Rendering by AttnBoost}\label{3.3}

During the second phase of the denoising cycle, as shown in Fig. \ref{fig:2}, we deploy the AttnBoost mechanism to achieve fine-grained text-guided rendering. This module strategically enhances text-relevant regions in the latent space by processing cross-attention maps from the double-stream transformer block. The query (\textit{Q}), key (\textit{K}), and value (\textit{V}) matrices are derived from the concatenation of the edited latent representation $\mathbf{z}_{\text{edit}}$ and the target text embeddings $\mathbf{e}^{\text{tar}}_p$, followed by linear projections through the transformer layers. This ensures precise semantic alignment with target descriptions while maintaining visual consistency with the source image structure.

Our attention computation begins with the standard scaled dot-product formulation:
\begin{equation}
\text{Attention}(Q,K,V) = \text{softmax}\left(\frac{QK^T}{\sqrt{d_k}} \right)V,
\end{equation}

\noindent\textbf{Text Region Enhancement} applies targeted amplification to text regions through element-wise transformation:
\begin{equation}
A_{\text{enhanced}}(b,h,q,k) = 
\begin{cases} 
\mathcal{T}(A(b,h,q,k)) & \text{if } q \in [\text{start}_1, \text{end}_1], \\
A(b,h,q,k) & \text{otherwise},
\end{cases}
\end{equation}
where $A \in \mathbb{R}^{B \times H \times L \times S}$ denotes the original attention tensor with batch size $B$, attention heads $H$, query length $L$, and key sequence length $S$. The transformation function $\mathcal{T}: \mathbb{R} \rightarrow \mathbb{R}$ implements the region-specific amplification.

\noindent\textbf{Attention Mapping and Aggregation} extracts text-to-image attention patterns and consolidates them through dimensional reduction:
\begin{align}
A_{\text{t2i}} &= A_{\text{enhanced}}[\cdot, \cdot, \mathcal{I}_{\text{text}}, \mathcal{I}_{\text{image}}], \\
A_{\text{agg}} &= \sum_{q \in \mathcal{I}_{\text{text}}} A_{\text{t2i}}[\cdot, \cdot, q, \cdot],
\end{align}
where $\mathcal{I}_{\text{text}} = [\text{start}_1, \text{end}_1]$ represents the text token indices, $\mathcal{I}_{\text{image}} = [N_{\text{text}}, S]$ denotes the image token indices, and $N_{\text{text}}$ indicates the quantity of text tokens in the input token sequence.

The extracted attention maps are further refined through spatial pooling, enabling the aggregation of local features and enhancing the focus on relevant regions:
\begin{equation}
\bar{A} = \frac{1}{B \times H \times W} \sum_{i=1}^{B} \sum_{j=1}^{H} \sum_{k=1}^{W} A_{i,j,k},
\end{equation}
where $\bar{A}$ represents the spatially pooled attention map, obtained by averaging the original attention tensor $A$ across batch, height, and width dimensions, with $W$ denoting the feature map width.

Normalization is then applied to ensure consistent value ranges and enhance numerical stability:
\begin{equation}
\hat{A} = \frac{\bar{A} - \min(\bar{A})}{\max(\bar{A}) - \min(\bar{A}) + \epsilon}, \quad \epsilon = 1 \times 10^{-8},
\end{equation}
where $\hat{A}$ denotes the normalized attention map constrained to $[0,1]$ range, while $\epsilon$ provides numerical stability to prevent division by zero.

The refined attention guidance is integrated into the denoising process through scheduler modulation:
\begin{equation}
z_{t-1} = \mathcal{S}(z_t, \hat{A}, t),
\end{equation}
where $z_t$ and $z_{t-1}$ represent the latent representations at current and subsequent timesteps, while $\mathcal{S}$ indicates the modified scheduler function that incorporates attention guidance at denoising step $t$. Further details regarding the $\mathcal{S}$ scheduler and its control enhancement through $\hat{A}$ will be elaborated in the Appendix.

AttnBoost establishes a mathematically grounded framework for transforming cross-modal attention patterns into spatial guidance signals. This systematic processing pipeline, from targeted region enhancement through normalized spatial guidance, enables precise text-controlled rendering while preserving structural integrity, providing a robust foundation for semantically aware image editing in complex visual environments.

\begin{table*}[t]
  \centering
  \small
  \caption{Performance comparison of different methods on the ScenePair dataset.}
  \vspace{-10px}
  \label{tab:1}
  \begin{tabular*}{\textwidth}{@{}c@{\extracolsep{\fill}}cccccc}
    \toprule
    \multirow{2}{*}{Methods} & \multicolumn{6}{c}{ScenePair} \\
    \cmidrule(lr){2-7}
    & SSIM ($\times10^{-2}$) $\uparrow$ & PSNR $\uparrow$ & MSE ($\times10^{-2}$) $\downarrow$ & FID $\downarrow$ & ACC (\%) $\uparrow$ & NED $\uparrow$ \\
    \midrule
    DiffSTE~\cite{ji2023improving} & 22.76 & 12.26 & 7.34 & 180.15 & 71.11 & 0.907 \\
    TextDiffuser~\cite{chen2023textdiffuserdiffusionmodelstext} & 26.99 & 13.93 & 5.70 & 56.67 & 51.48 & 0.719 \\
    AnyText~\cite{tuo2023anytext} & 30.73 & 13.66 & 6.05 & 51.44 & 51.12 & 0.734 \\
    TextFlux~\cite{xie2025textflux} & 86.57 & 17.96 & 1.83 & 54.64 & \textbf{80.40} & 0.911 \\
    Flux-fill~\cite{flux2024} & 82.73 & 17.10 & 2.99 & 107.83 & 13.74 & 0.306 \\
    Flux-Kontext~\cite{labs2025flux1kontextflowmatching} & 87.08 & 20.53 & 1.58 & \underline{15.41} & 78.72 & \textbf{0.920} \\
    Qwen-image~\cite{wu2025qwen} & 77.89 & 15.14 & 4.19 & 56.71 & 68.59 & 0.833 \\
    FlowEdit~\cite{kulikov2024flowedit} & \underline{87.60} & \underline{20.89} & \underline{1.16} & 25.41 & 45.51 & 0.590 \\
    \midrule
    TextFlow (Ours) &\textbf{89.03} & \textbf{22.47} &\textbf{ 0.91} & \textbf{13.53} & \underline{79.98} & \underline{0.914} \\
    \bottomrule
  \end{tabular*}
\end{table*}



\section{Experiments}
\label{sec:exper}

        

\begin{figure*}[t]
    \centering
    \includegraphics[width=1\linewidth]{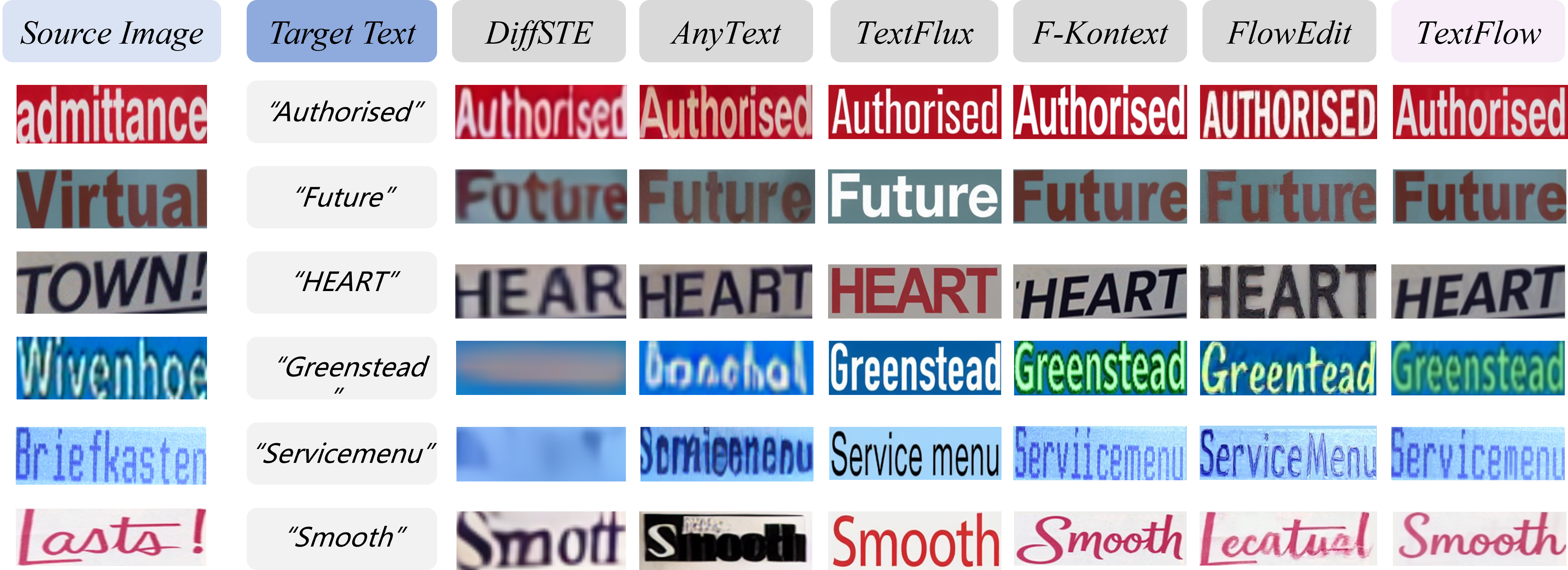}
    \caption{Qualitative Analysis. The compared methods include both training-based STE approaches like DiffSTE~\cite{ji2023improving}, AnyText~\cite{tuo2023anytext}, TextFlux~\cite{xie2025textflux} and recent training-free editing techniques FlowEdit~\cite{kulikov2024flowedit}. We also include the powerful foundational model Flux-Kontext~\cite{labs2025flux1kontextflowmatching} (F-Kontext),  for a more extensive comparison.}
    \vspace{-10px}
    \label{fig:4}
\end{figure*}

\subsection{Datasets and metrics}

\textbf{Datasets.} 
To provide assessments on both image generation quality and visual text quality, we employ the ScenePair dataset~\cite{zeng2024textctrl}, a real-world scene text image-pair dataset. Specifically, ScenePair comprises 1,280 image pairs with text labels sourced from ICDAR 2013~\cite{karatzas2013icdar}, HierText~\cite{long2023icdar}, and MLT 2017~\cite{2017ICDAR2017}. Each pair consists of two cropped text images that share similar text length, style, and background, along with the corresponding original full-size images. To ensure consistent input dimensions across all models, we pad the cropped images with background-similar colors to a resolution of 384×256, and all metrics are computed based on this preprocessed input.

\noindent\textbf{Evaluation Metrics.} 
For the assessment of image generation quality, we employ the following metrics: (1) Structural Similarity Index Measure (SSIM): Measures the structural similarity between the generated image and the Ground Truth (GT); (2) Peak Signal-to-Noise Ratio (PSNR): calculate the peak signal-to-noise ratio to assess the distortion level by computing the mean squared error between the generated image and the GT; (3) Mean Squared Error (MSE): Quantifies the pixel-wise difference between the generated image and the GT; (4) Fréchet Inception Distance (FID): Evaluates the quality of synthesized images by comparing the statistical distributions of feature embeddings from the generated and GT images. For visual text quality assessment, we utilize Accuracy (ACC) and Normalized Edit Distance (NED) \cite{heusel2017gans} to evaluate the correctness and overall quality of the generated text image, using an official text recognition algorithm \cite{baek2019wrongscenetextrecognition} and the corresponding checkpoint.
\subsection{Implementation Details}
Our proposed TextFlow framework is built upon the FLUX-Kontext \cite{labs2025flux1kontextflowmatching} model as the core image editing generator due to its superior performance in generating high-quality images. For the text encoder, we utilize the T5 and CLIP to extract text embeddings, which provide a robust semantic representation for both the source and target prompts. The entire framework operates in a training-free manner, and no components are fine-tuned on any scene text editing datasets.
During the inference process, we employ the Overshoot \cite{hu2025amo} and Euler scheduler with 50 denoising steps to balance generation quality and computational efficiency. 
All experiments are performed on a server equipped with 4 NVIDIA A6000 GPUs with 48G VRAM each. Additional experimental settings and implementation details will be provided in the Appendix.
\subsection{Comparison with State-of-the-Art Methods}
\textbf{Quantitative Analysis.}
We conduct a comprehensive evaluation of our proposed TextFlow framework against state-of-the-art methods on the ScenePair dataset. As summarized in Table~\ref{tab:1}, the compared methods include both training-based STE approaches like DiffSTE~\cite{ji2023improving}, TextDiffuser~\cite{chen2023textdiffuserdiffusionmodelstext}, AnyText~\cite{tuo2023anytext}, TextFlux~\cite{xie2025textflux} and recent training-free editing techniques FlowEdit~\cite{kulikov2024flowedit}. We also include the powerful foundational model Flux-fill~\cite{flux2024}, Flux-Kontext~\cite{labs2025flux1kontextflowmatching}, and Qwen-image~\cite{wu2025qwen} for a more extensive comparison.

The experimental results demonstrate the superior performance of our method across multiple dimensions. In terms of image quality and structural fidelity, our approach achieves the highest SSIM score of 89.03 and the best PSNR of 22.47, significantly outperforming all competing methods. Notably, our method reduces the MSE to 0.91, approximately 42\% lower than the second-best method, Flux-Kontext~\cite{labs2025flux1kontextflowmatching}, indicating superior pixel-level reconstruction accuracy. The lowest FID score of 13.53 further confirms that our generated images are statistically closest to the real data distribution, highlighting exceptional visual realism.

Regarding textual rendering accuracy, our method achieves a competitive character-level accuracy of 79.98\% and NED score of 0.914. While TextFlux~\cite{xie2025textflux} shows a slightly higher accuracy of 80.40\%, our method maintains a better balance between textual correctness and visual quality, as evidenced by our substantially superior FID and PSNR metrics. This balanced performance is practically crucial for real-world applications where both textual accuracy and visual coherence are paramount. A comprehensive experimental evaluation of additional methods will be provided in the Appendix.

\noindent\textbf{Qualitative Analysis.}
Fig.~\ref{fig:4} presents a qualitative comparison of generated results. Our proposed TextFlow is evaluated against several representative methods, including UNet-based approaches such as DiffSTE~\cite{ji2023improving} and AnyText~\cite{tuo2023anytext}, as well as state-of-the-art DiT-based methods in STE like TextFlux~\cite{xie2025textflux}, FLUX-Kontext~\cite{labs2025flux1kontextflowmatching}, and FlowEdit~\cite{kulikov2024flowedit}. For methods requiring mask-conditioned inputs, such as AnyText~\cite{tuo2023anytext} and TextFlux~\cite{xie2025textflux}, we applied background-colored padding to the input images to maintain consistent input resolution. Regarding prompt design, the source description was uniformly formatted as: \textit{“A picture with word ‘{$T_{src}$}’.”}, while the target prompt followed the structured template: \textit{“Please replace the word ‘{$T_{src}$}’ with ‘{$T_{tar}$}’.”}.

While TextFlux~\cite{xie2025textflux} maintains relatively high text accuracy, it suffers from significant style loss. Conversely, FLUX-Kontext~\cite{labs2025flux1kontextflowmatching} demonstrates better style preservation but shows deficiencies in text accuracy. FlowEdit~\cite{kulikov2024flowedit}, as a training-free approach, achieves reasonable performance in both style consistency and text accuracy, yet falls short in handling fine-grained details such as letter case consistency and glyph structure. In contrast, as demonstrated in the fifth row with the word “\textit{Servicemenu}” and the sixth row with “\textit{Smooth}”, our method achieves superior performance in both style preservation and text accuracy while maintaining excellent detail handling capabilities.

\begin{figure}[!htbp]
    \centering
    \includegraphics[width=1\linewidth]{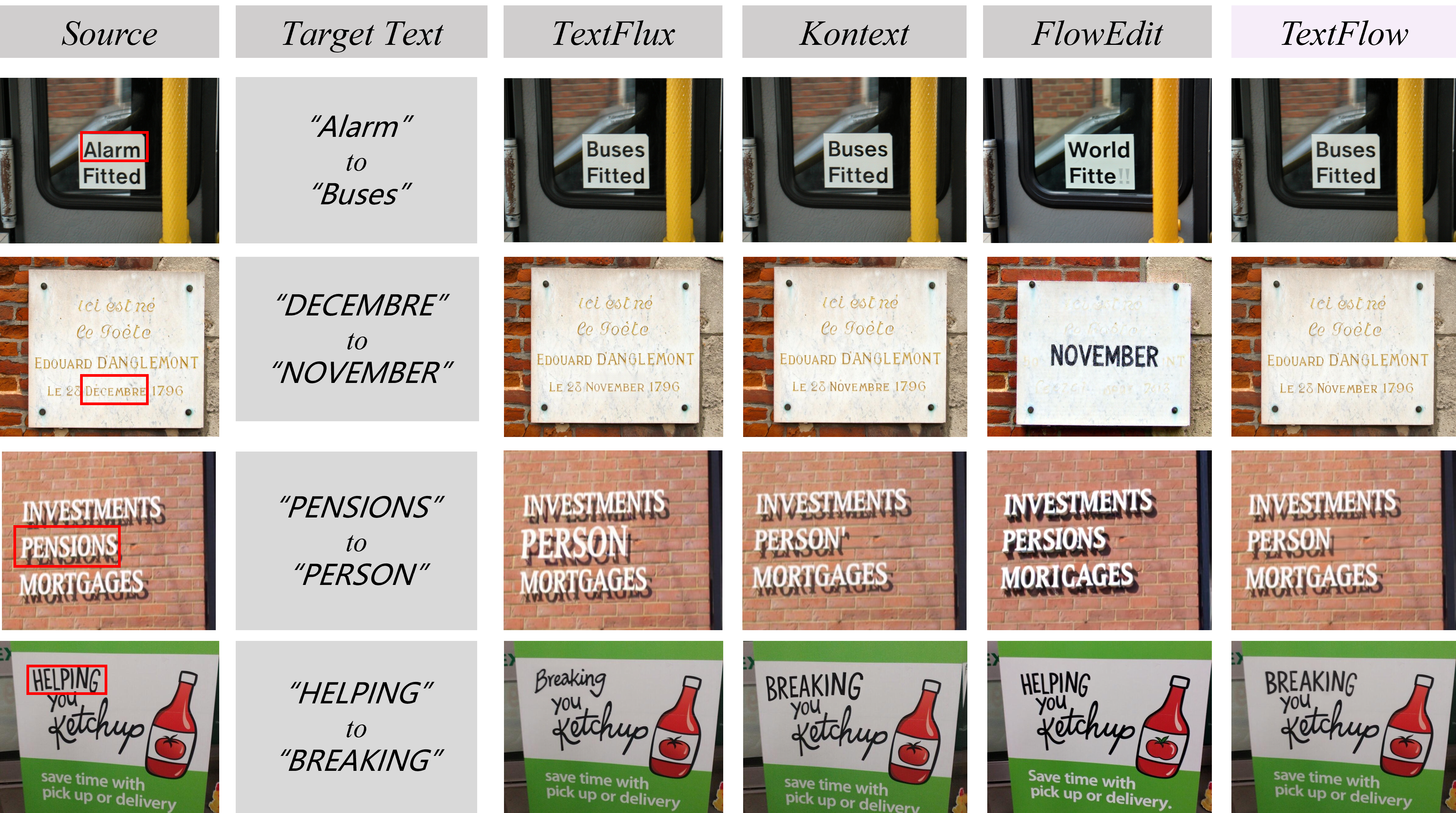}
    \vspace{-10px}
    \caption{Qualitative comparison among different DiT-based methods on a full-size image. }
    \vspace{-10px}
    \label{fig:5}
\end{figure}

Fig.~\ref{fig:5} shows editing results on full-size images, where TextFlow achieves competitive performance in style preservation and text accuracy against other DiT-based methods, underscoring its superior editing capability.

\subsection{Ablation Study}
To comprehensively evaluate the contributions of different components in our proposed framework, we conduct systematic ablation studies across three key aspects: the FMS module for structural preservation, the AttnBoost mechanism for text rendering accuracy, and the optimization of inference configurations, including scheduler selection and step count. These experiments validate the necessity of each component and identify optimal parameter settings.

\begin{table}[!htbp]
\centering
\small
\caption{Ablation of FMS modules for image quality.}
\vspace{-10px}
\begin{tabular}{lcccc}
\toprule
FMS Module & SSIM $\uparrow$ & PSNR $\uparrow$ & MSE $\downarrow$ & FID $\downarrow$ \\
\midrule
FlowEdit~\cite{kulikov2024flowedit} & \underline{87.60} & \underline{20.89} & \underline{1.16} & 25.41\\
Ours w/o FMS & 87.09 & 20.47 & 1.35 & \underline{16.69} \\
Ours w FMS & \textbf{89.04} & \textbf{22.42} & \textbf{0.97} & \textbf{13.52} \\
\bottomrule
\end{tabular}
\label{tab:2}
\end{table}

Table \ref{tab:2} presents the ablation results evaluating our proposed FMS module. Our full method with FMS achieves the best performance across all image quality metrics, with 89.04 SSIM, 22.42 PSNR, 0.97 MSE, and 13.52 FID.

Compared to FlowEdit~\cite{kulikov2024flowedit}, our method shows substantial improvements, increasing SSIM from 87.60 to 89.04 and PSNR from 20.89 to 22.42 while reducing FID from 25.41 to 13.52. Removing the FMS module causes significant degradation, with PSNR dropping by 1.95 and MSE increasing by 39.2\%, confirming the critical importance of our trajectory correction. Although the ablated version maintains an FID advantage over FlowEdit~\cite{kulikov2024flowedit}, the comprehensive superiority of our full method demonstrates that FMS effectively balances structural preservation with visual quality enhancement.

\begin{figure}[!htbp]
    \centering
    \includegraphics[width=1.0\linewidth]{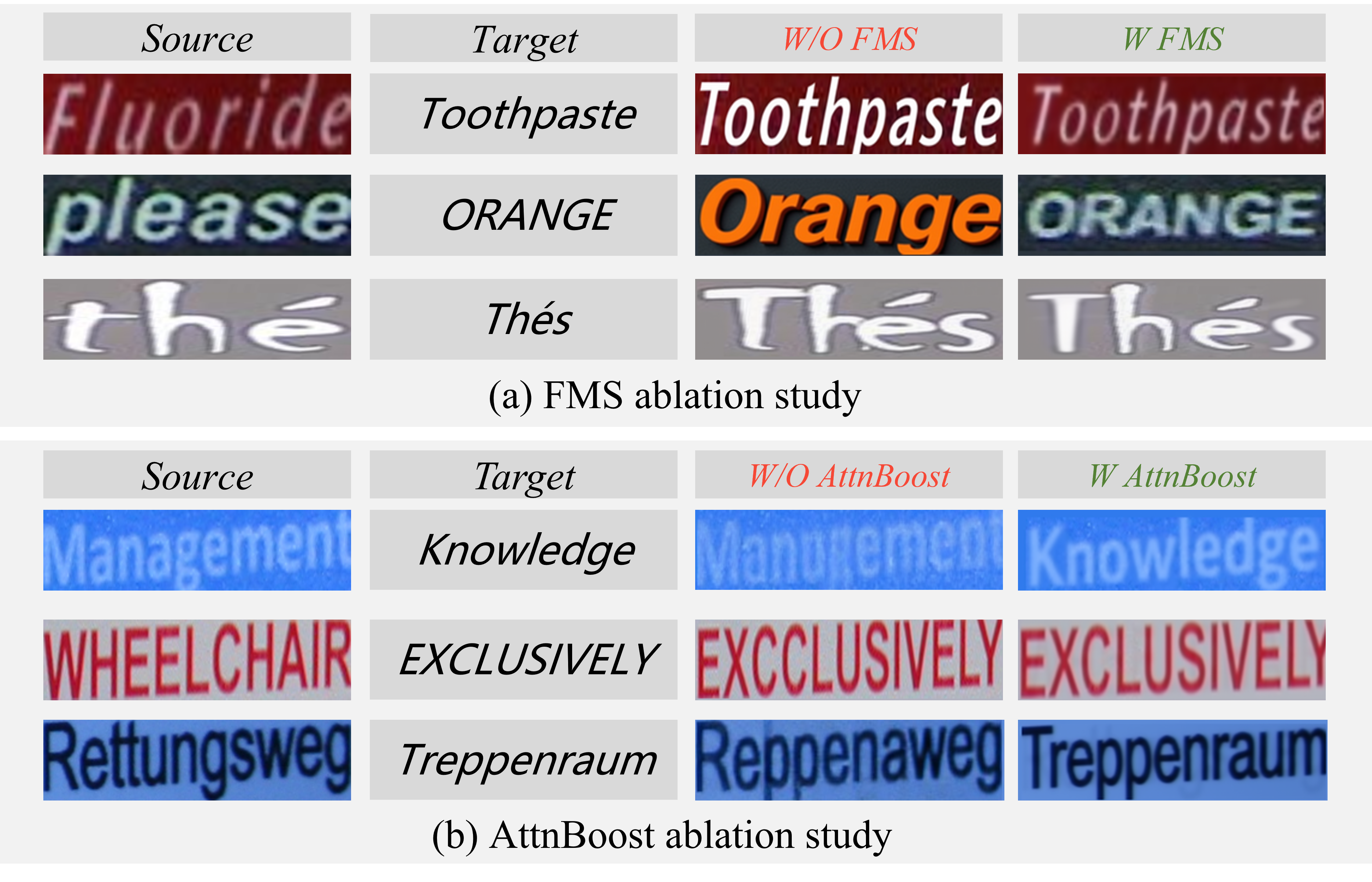}
    \vspace{-15px}
    \caption{Qualitative ablation studies validate the effectiveness of FMS in style preservation and demonstrate the significant improvement in text rendering accuracy achieved by AttnBoost.}
    \vspace{-10px}
    \label{fig:6}
\end{figure}

As demonstrated in Fig.~\ref{fig:6} (a), the incorporation of FMS significantly enhances style consistency between the original and edited images while notably improving the preservation of fine-grained details.


\begin{table}[!htbp]
\centering
\footnotesize
\vspace{-5px}
\caption{Ablation of AttnBoost considering text accuracy.}
\vspace{-10px}
\begin{tabular}{lcccc}
\toprule
\multirow{2}{*}{AttnBoost   Module} & \multicolumn{2}{c}{ScenePair} & \multicolumn{2}{c}{ScenePair (Random)} \\ \cline{2-5} 
& ACC(\%) $\uparrow$ & NED $\uparrow$ & ACC(\%) $\uparrow$ & NED $\uparrow$  \\ 
\midrule
FLUX-Kontext~\cite{labs2025flux1kontextflowmatching} & \underline{78.72} & \underline{0.920} & \textbf{76.63} & \textbf{0.916} \\ 
Ours w/o AttnBoost & 20.35 & 0.420 & 18.84 & 0.391 \\
Ours w AttnBoost & \textbf{79.80} & \textbf{0.931} & \underline{74.52} & \underline{0.874} \\ 
\bottomrule
\end{tabular}
\label{tab:3}
\end{table}

Table~\ref{tab:3} presents that the AttnBoost module can significantly enhance textual accuracy. On the ScenePair dataset, our full model with AttnBoost achieves the best performance with 79.80\% accuracy and 0.931 NED, outperforming both the FLUX-Kontext~\cite{labs2025flux1kontextflowmatching} baseline and the ablated version. Although FLUX-Kontext~\cite{labs2025flux1kontextflowmatching} performs best on the more challenging ScenePair Random dataset, our method remains competitive. Removing AttnBoost causes a dramatic performance drop, with accuracy decreasing by approximately 75\% and NED by 55\%, confirming its essential role in high-quality text rendering.

The Fig.~\ref{fig:6} (b) reveals that AttnBoost substantially improves textual accuracy, with particularly notable enhancements observed in challenging cases involving long words and consecutive characters.

\begin{table}[!htbp]
\centering
\footnotesize
\vspace{-5px}
\caption{Ablation of inference steps on ScenePair.}
\vspace{-10px}
\begin{tabular}{@{}lcccccc@{}}
\toprule
Steps & SSIM $\uparrow$ & PSNR $\uparrow$ & MSE $\downarrow$ & FID $\downarrow$ & ACC(\%) $\uparrow$ & NED $\uparrow$  \\ 
\midrule
24 & 86.80 & 20.21 & 1.43 & 16.94 & 77.97 & \underline{0.925}  \\
30 & 87.12 & 19.86 & 1.46 & 23.1 & \underline{79.90} &  \textbf{0.928} \\ 
42 & \underline{88.04} & \underline{22.21} & 0.97 & 52.8 & 79.40 & 0.926 \\
50 & \textbf{89.30} & \textbf{22.47} & \underline{0.91} & \underline{13.53} & \textbf{79.98} & 0.914 \\
70 & 87.01 & 21.02 &\textbf{0.90} & \textbf{12.83} & 79.88 & 0.914 \\
\bottomrule
\end{tabular}
\label{tab:4}
\end{table}

Table~\ref{tab:4} presents a comprehensive comparison of inference steps across both generative and render metrics. Our experiments demonstrate that 50 denoising steps achieve the optimal balance between generation quality and textual accuracy while maintaining computational efficiency.

In terms of image quality metrics, 50 steps yield the best overall performance with 89.30 SSIM, 22.47 PSNR, and 13.53 FID, while achieving a competitive MSE of 0.91. For textual accuracy, 50 steps produce the highest character accuracy of 79.98\% with 0.914 NED. Although 70 steps achieve slightly better MSE and FID scores, the improvements are marginal while requiring significantly more computational resources.

The results indicate that 50 steps yield the most efficient operating point, delivering superior visual quality and text fidelity without the computational overhead associated with higher step counts. This balanced performance makes 50 steps the recommended setting for practical applications where both quality and efficiency are prioritized.

\begin{table}[!htbp]
\centering
\footnotesize
\caption{Ablation of scheduler on the ScenePair dataset.}
\vspace{-10px}
\begin{tabular}{lcc}
\toprule
Scheduler & ACC(\%) $\uparrow$ & NED $\uparrow$  \\ 
\midrule
Ours w Euler & 78.73 & 0.920  \\
Ours w Overshoot \cite{hu2025amo} & \textbf{79.90} & \textbf{0.931}  \\ 
\bottomrule
\end{tabular}
\vspace{-10px}
\label{tab:5}
\end{table}

Table~\ref{tab:5} presents that the Overshoot scheduler consistently outperforms the Euler scheduler in text rendering accuracy. Our method with the Overshoot scheduler achieves superior performance, reaching 79.90\% accuracy and 0.931 NED, compared to 78.73\% accuracy and 0.920 NED with the Euler scheduler. This demonstrates that the Overshoot scheduler, which extends the denoising trajectory beyond conventional bounds, provides more precise control over text generation, thereby improving character accuracy and editing quality.

\section{Conclusion and Limitation}
\label{sec:con}

We introduce TextFlow, a training-free framework for scene text editing that balances structural preservation with textual accuracy. It integrates two complementary components: FMS maintains structural consistency via trajectory guidance in early phases, while AttnBoost enables fine-grained text rendering in later phases. This integration establishes a new paradigm for phase-aware generative guidance. Extensive experiments demonstrate state-of-the-art performance in both image quality and text accuracy, delivering high-fidelity edits without task-specific training or large-scale paired datasets.

Despite these advances, certain limitations remain. The computational overhead of the underlying diffusion model limits real-time applicability, especially for high-resolution outputs. More notably, the framework struggles with multi-line text and complex layouts, where maintaining spatial and typographic consistency proves challenging.


\section*{Acknowledgement}
This work is Funded by Basic Research Program of Jiangsu (BK20251441, BK20252040, BK20251414).
{
    \small
    \bibliographystyle{ieeenat_fullname}
    \bibliography{main}
}
\clearpage
\setcounter{page}{1}
\maketitlesupplementary

\section{AttnBoost Mechanism and Overshoot Scheduler }
\label{sec:rationale}

\subsection{Attention-Modulated Overshooting}

The AttnBoost module integrates attention mechanisms to achieve adaptive control over overshooting intensity, specifically targeting text regions while preserving non-text areas.

\begin{figure}[h]
    \centering
    \includegraphics[width=0.8\linewidth]{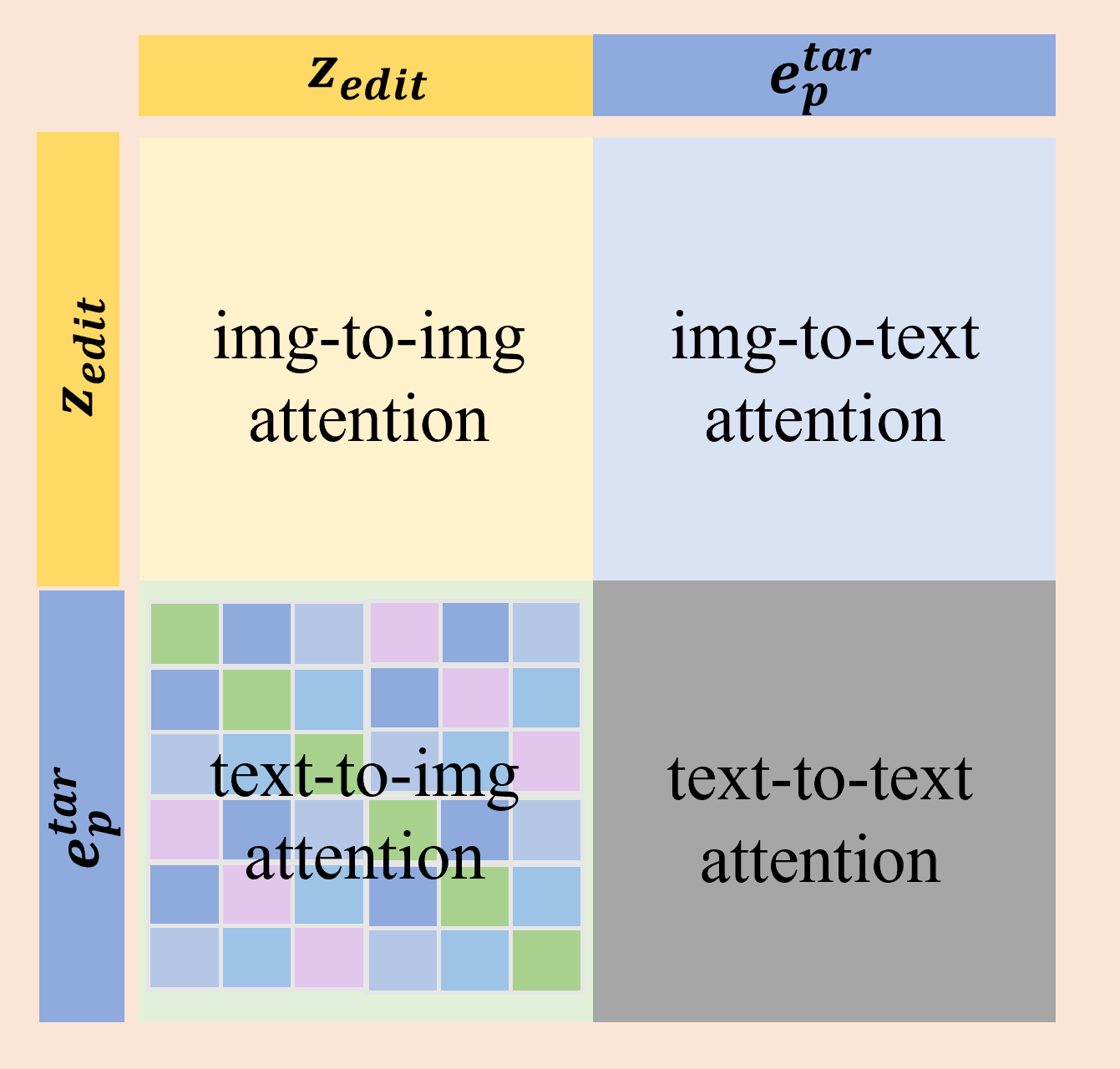}
    \caption{Computation graph of the full attention map between $\mathbf{z}_{\text{edit}}$ and $\mathbf{e}^{\text{tar}}_p$. The green-highlighted region in the lower-left corner illustrates the text-to-image attention components ($A_{\text{t2i}}$) extracted from each DoubleStream Transformer block, which are subsequently utilized for scheduler enhancement in the text rendering.}
    \label{fig:7}
\end{figure}

As shown in Fig.~\ref{fig:7}, attention mapping and aggregation mentioned in Sec.~\ref{3.3} extract text-to-image attention patterns and consolidate them through dimensional reduction. The resulting attention maps are then refined via spatial pooling to concentrate relevant information, followed by normalization to ensure consistent value ranges and numerical stability.

\subsection{Implementation of Overshoot Scheduler}

\begin{figure}[h]
    \centering
    \includegraphics[width=1\linewidth]{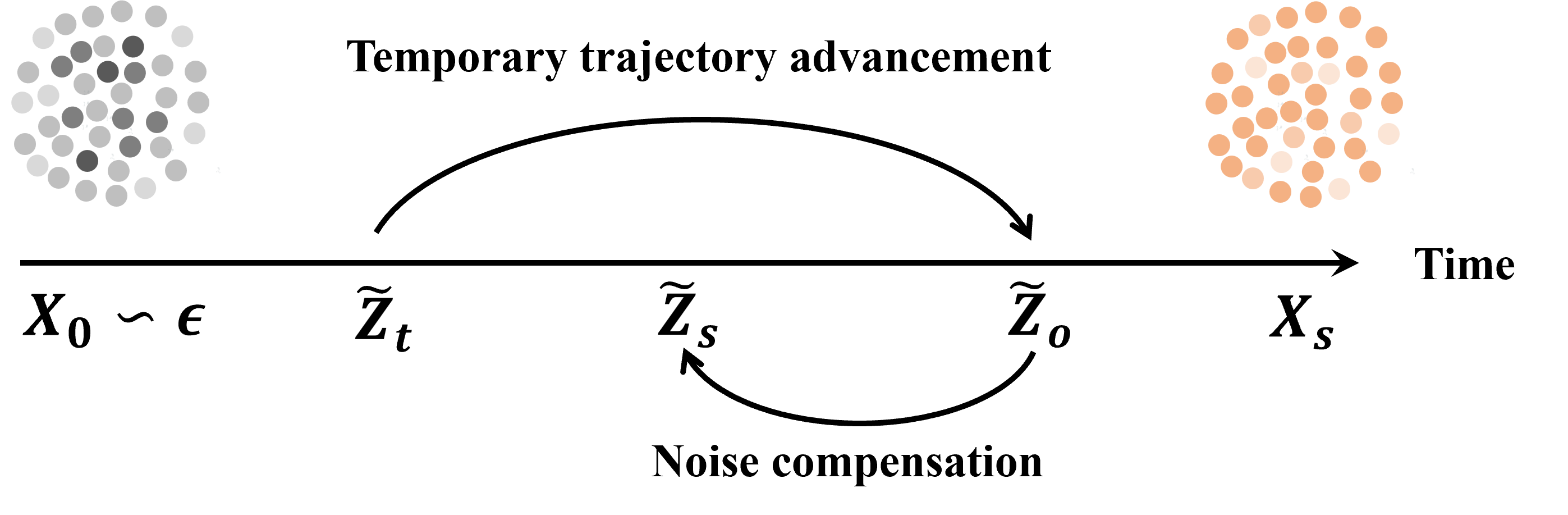}
    \caption{Schematic diagram of the Overshoot scheduler. Given the latent representation $\tilde{Z}_t$ at time $t$, the scheduler first advances the learned ODE trajectory beyond the target step to obtain $\hat{Z}_o$, then applies calibrated noise to return to the corrected state $\tilde{Z}_s$. The injected noise is precisely controlled to ensure $\tilde{Z}_s$ conforms to the marginal distribution of $X_s$.}
    \vspace{-10px}
    \label{fig:8}
\end{figure}

The Overshoot scheduler \cite{hu2025amo} implements a controlled trajectory deviation mechanism during the diffusion sampling process, leveraging attention-guided overshooting to enhance text rendering fidelity. The process, shown in Fig.~\ref{fig:8}, begins with a sample from the initial noise distribution, \(\tilde{Z}_0 = X_0 \sim \pi_0\), and aims to compute the latent representation \(\tilde{Z}_s\) at time \(s = t + \varepsilon\) from the current state \(\tilde{Z}_t\), where \(\varepsilon > 0\) is the denoising step size.

In contrast to the standard Euler sampler, which updates as \(\tilde{Z}_s = \tilde{Z}_t + \varepsilon v_\theta(\tilde{Z}_t, t)\), our overshooting sampler incorporates stochastic noise and attention modulation through a two-step procedure:

\begin{enumerate}
    \item \textbf{Temporary trajectory advancement:}
    The sampler first advances from the current timestep \(t\) to an overshoot point \(o = s + \varepsilon c \hat{A}\), where \(c \in \mathbb{R}^+ \) is the overshoot intensity parameter and \(\hat{A}\) denotes the normalized attention map derived from cross-modal interactions. The advanced latent representation is computed as:
    \begin{align}
    \hat{Z}_o &= \tilde{Z}_t + v_\theta(\tilde{Z}_t, t) \odot (o - t) \notag \\
             &= \tilde{Z}_t + \varepsilon (1 + c \hat{A}) \odot v_\theta(\tilde{Z}_t, t),
    \end{align}
    Here, \(v_\theta(\tilde{Z}_t, t)\) represents the velocity field parameterized by a neural network, and \(\odot\) denotes element-wise multiplication with the attention map \(\hat{A}\).

    \item \textbf{Noise compensation and trajectory correction:}
    The oversampled latent \(\hat{Z}_o\) is then corrected back to the target time \(s\) by introducing stochastic noise:
    \begin{equation}
        \tilde{Z}_s = a \hat{Z}_o + b \xi, \quad \xi \sim \mathcal{N}(0, I).
    \end{equation}
    The correction coefficients \(a\) and \(b\) are defined as:
    \begin{align}
        a &= \frac{s}{o}, \\
        b &= \sqrt{(1 - s)^2 - \frac{s^2 (1 - o)^2}{o^2}},
    \end{align}
    This step ensures stability by compensating for the overshooting effect while preserving textual details.
\end{enumerate}

The overall scheduler output for the next timestep is thus given by:
\begin{equation}
    z_{t-1} = \mathcal{S}(z_t, \hat{A}, t),
\end{equation}
where \(\mathcal{S}\) encapsulates the overshooting and correction steps. This approach enables targeted improvements in text rendering quality within attention-masked regions without full model fine-tuning, relying on well-aligned attention maps for optimal performance. The integration of attention modulation allows for adaptive control over overshooting intensity, focusing on text-relevant areas while minimizing artifacts in non-text regions.

\begin{table*}[h]
  \centering
  \small
  \caption{Performance of more methods on the ScenePair dataset.}
  \vspace{-10px}
  \label{tab:6}
  \begin{tabular*}{\textwidth}{@{}c@{\extracolsep{\fill}}cccccc}
    \toprule
    \multirow{2}{*}{Method} & \multicolumn{6}{c}{ScenePair} \\
    \cmidrule(lr){2-7}
    & SSIM ($\times10^{-2}$) $\uparrow$ & PSNR $\uparrow$ & MSE ($\times10^{-2}$) $\downarrow$ & FID $\downarrow$ & ACC (\%) $\uparrow$ & NED $\uparrow$ \\
    \midrule
    TextCtrl~\cite{zeng2024textctrl} & 37.56 & 14.99 & 4.47 & 43.78 & \textbf{84.67} & \textbf{0.936 }\\
     Flux-Text~\cite{lan2025flux} & 86.45 & 17.95 & 1.93 & 54.84 & 70.94 & 0.877 \\
    TextFlow (Ours) &\textbf{ 89.03} & \textbf{22.47} &\textbf{ 0.91} & \textbf{13.53} & \underline{79.98} & \underline{0.914} \\
    \bottomrule
  \end{tabular*}
\end{table*}

\begin{table*}[h]
\centering
\footnotesize
\caption{Performance of different methods on TamperScene and AnyText-Bench. HE represents Human Evaluation.}
\vspace{-10px}
\scalebox{1}{
\begin{tabular}{@{}llccccccc@{}}
\toprule
\multirow{2}{*}{Type} & \multirow{2}{*}{Method} & ScenePair & \multicolumn{3}{c}{TamperScene-2k} & \multicolumn{3}{c}{AnyText-Bench-en} \\
\cmidrule(lr){3-3} \cmidrule(lr){4-6} \cmidrule(l){7-9}
& & HE & ACC(\%)  $ \uparrow $  & NED  $ \uparrow $  & HE & Sen.acc(\%) & NED & HE \\ 
\midrule
\multirow{5}{*}{Training-Based} 
& TextFlux~\cite{xie2025textflux}       & 6.9 & \underline{19.70} & 0.42 & 5.4 &  8.13  & 0.21  &   6.2     \\
    & Flux-Text~\cite{lan2025flux}      & 7.1 & 18.60 & 0.42 & 5.9 &  \textbf{38.89} & \textbf{0.65}  &   \underline{8.1}     \\
& Qwen-image~\cite{wu2025qwen} & 8.0 & 10.75 & 0.37 & 7.7 &  3,74  & 0.15  &   7.5     \\
& Flux-Kontext~\cite{labs2025flux1kontextflowmatching}   & 7.3 & 17.79 & 0.45 & 7.2 &  19.53 & 0.39  &   7.9     \\
& Longcat-Edit~\cite{LongCat-Image}   & 8.2 &  0.65 & 0.29 & 4.9 &  5.66  & 0.26  &   8.1     \\
\midrule
\multirow{3}{*}{Training-Free} 
& FlowEdit~\cite{kulikov2024flowedit}         & 7.5 &  5.56 & 0.25 & 6.9 & 1.60  & 0.10   &  5.3      \\
& TextFlow-Kontext & \underline{8.3} & 18.75 & \textbf{0.45} & \underline{8.0} & 26.07 & 0.45   &  8.0      \\
& TextFLow-Longcat & \textbf{8.5} & \textbf{20.95} & \underline{0.44} & \textbf{8.7} & \underline{38.75} & \underline{0.61}   &  \textbf{8.5}      \\
\bottomrule
\end{tabular}}
\label{tab:7}
\end{table*}

\section{More Analysis of Experiments }
In this section, we present additional experiments to comprehensively analyze and validate our method.
\subsection{Comparison with SOTA}
\noindent\textbf{Datasets.}
ScenePair collects 1,280 image pairs with text labels from ICDAR 2013~\cite{karatzas2013icdar}, HierText~\cite{long2023icdar}, and MLT 2017~\cite{2017ICDAR2017}, where each pair consists of two cropped text images with similar text length, style, and background, along with the original full-size images. We conduct quantitative analysis on the ScenePair dataset, where we pad the cropped images with similar background colors to a resolution of 384×256 to ensure consistent input size across all models, and all metrics are computed based on this preprocessing. We perform qualitative analysis using challenging full-size scene images selected from ICDAR 2013~\cite{karatzas2013icdar}, HierText~\cite{long2023icdar}, and MLT 2017~\cite{2017ICDAR2017} datasets.

\noindent\textbf{Quantitative Analysis.}
As shown in Table~\ref{tab:6}, our method achieves state-of-the-art performance across most image-quality metrics on the ScenePair dataset. TextFlow significantly outperforms competing methods in structural preservation with an SSIM of 89.03 and a PSNR of 22.47, while also demonstrating superior distortion reduction with an MSE of 0.91 and an FID of 13.53. Although TextCtrl attains the highest text accuracy with a character accuracy of 84.67\% and an NED of 0.936, our method maintains competitive textual performance with 79.98\% accuracy and 0.914 NED while delivering better overall visual quality and generalization capability. These quantitative results confirm TextFlow's balanced approach to preserving scene structure while achieving accurate text rendering.

Table~\ref{tab:7} presents an extensive quantitative comparison of different methods on the TamperScene and AnyText-Bench datasets. Among training-based approaches, methods such as Longcat-edit and Flux-text demonstrate strong performance across various metrics. In the training-free category, particularly TextFlow-Longcat, our proposed TextFlow variants achieve the highest human evaluation (HE) scores and competitive accuracy metrics, outperforming existing training-free methods and narrowing the gap with training-based approaches. These results validate the effectiveness of our phase-aware guidance strategy in preserving style and ensuring textual accuracy without requiring task-specific training.

\begin{figure*}[h]
    \centering
    \includegraphics[width=1\linewidth]{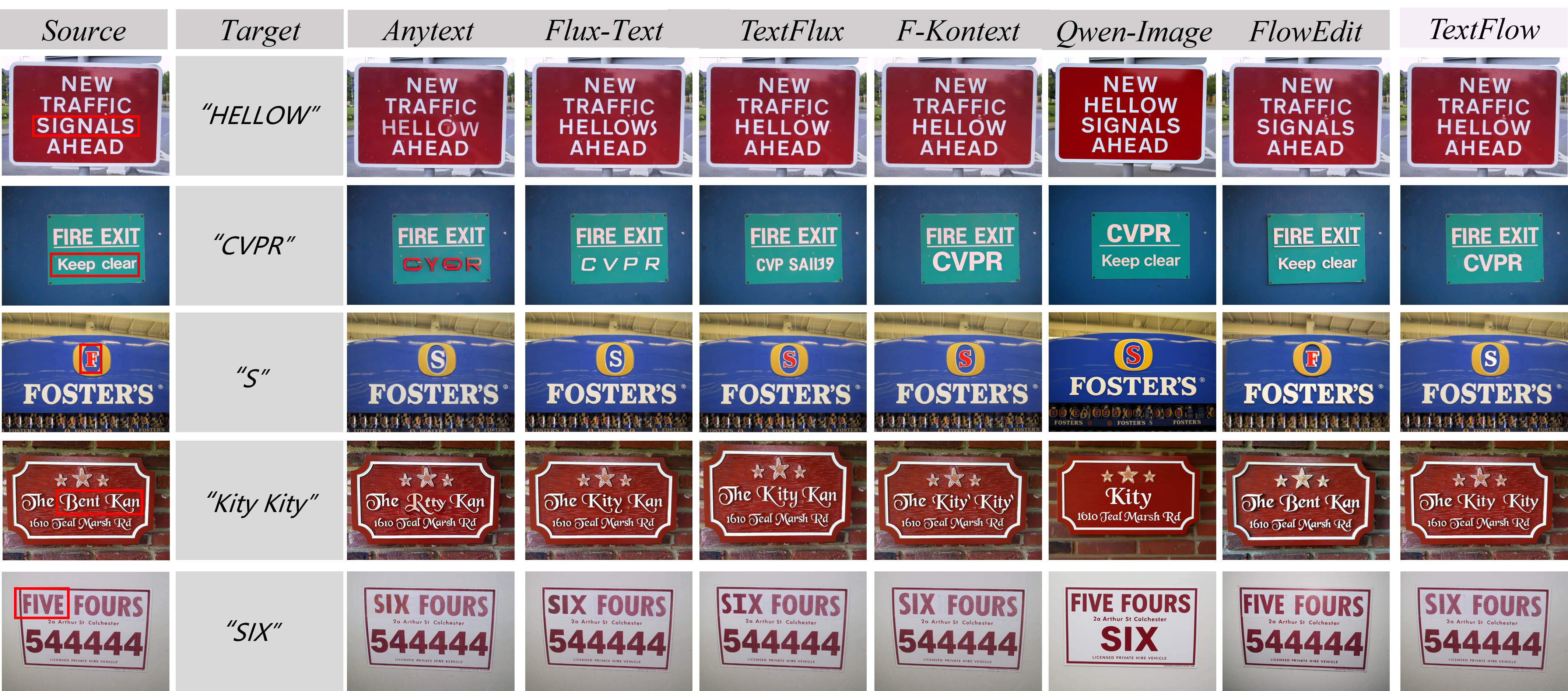}
    \caption{Comparative analysis of editing performance with additional models on full-size images.}
    \label{fig:10}
\end{figure*}

\noindent\textbf{Qualitative Analysis.} For comprehensive comparison on full-size images, as shown in Fig.~\ref{fig:10}, we select representative methods including: AnyText~\cite{tuo2023anytext} from UNet-based approaches; Flux-Text~\cite{lan2025flux} and textFlux~\cite{xie2025textflux} from DiT-based STE specific methods; general editing models Flux-Kontext~\cite{labs2025flux1kontextflowmatching} and Qwen-image~\cite{wu2025qwen}; and the training-free approach FlowEdit~\cite{kulikov2024flowedit}. The results demonstrate that our method achieves superior performance in both style consistency and text accuracy. Notably, our approach successfully renders challenging out-of-vocabulary words like “HELLOW” in the first example, demonstrating its robust semantic understanding and effective visual-textual alignment. Furthermore, the integrated attention mechanism enables precise spatial localization of target regions without requiring explicit masks, achieving accurate local editing through attention-guided refinement in a fully mask-free paradigm.

\subsection{Strength of $V_{\Delta}$ in FMS}
Table~\ref{tab:8} presents an ablation study on the strength parameter \(V_{\Delta}\) using the ScenePair dataset. As the strength increases from 0.2 to 1.0, most metrics improve, with SSIM, PSNR, and ACC reaching their highest values at strength 1.0 or 2.0, while FID achieves the lowest (best) at strength 5.0. Notably, a strength of 1.0 yields a balanced performance with high SSIM (89.03), PSNR (22.47), ACC (79.98\%), and competitive FID (13.53). Beyond 1.0, although image similarity metrics (SSIM, PSNR) continue to improve slightly, textual accuracy (ACC, NED) begins to decline, indicating a trade-off between style preservation and text fidelity. These results suggest that a moderate strength of around 1.0 optimally balances the two objectives.

\begin{table}[t]
\centering
\footnotesize
\caption{Ablation of $V_{\Delta}$ strength on ScenePair.}
\vspace{-10px}
\scalebox{1}{\begin{tabular}{@{}lcccccc@{}}
\toprule
Stren. & SSIM $\uparrow$ & PSNR $\uparrow$ & MSE $\downarrow$ & FID $\downarrow$ & ACC(\%) $\uparrow$ & NED $\uparrow$  \\ 
\midrule
0.2 & 84.81 & 18.19 & 2.67 & 23.16 & 75.20 & 0.895  \\
0.5 & 85.14 & 19.83 & 2.06 & 19.21 & 76.92 &  0.908 \\
0.7 & 85.92 & 20.86 & 1.46 & 17.25 & \underline{79.90} &  \textbf{0.928} \\
1.0 & \underline{89.03} & \underline{22.47} & \underline{0.91} & \textbf{13.53} & \textbf{79.98} & \underline{0.914} \\
2.0 & \textbf{89.30} & \textbf{23.47} & \textbf{0.90} & \underline{14.20} & 77.62 & 0.894 \\
5.0 & 88.47 & 21.02 & 0.90 & \textbf{12.83} & 76.88 & 0.872 \\
\bottomrule
\end{tabular}}
\label{tab:8}
\end{table}

\subsection{Strength of AttnBoost}

\begin{figure}
    \centering
    \includegraphics[width=1\linewidth]{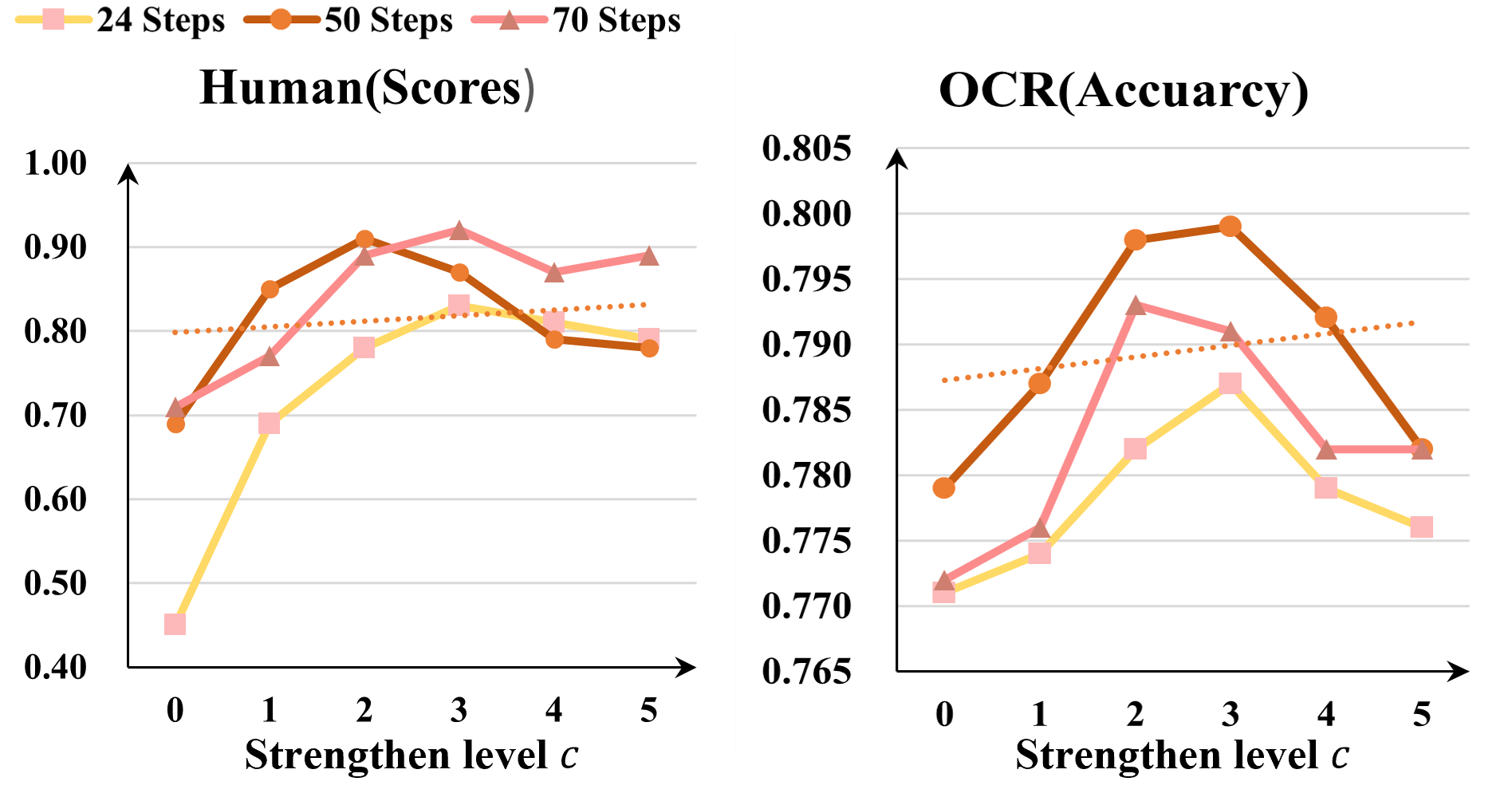}
    \caption{Ablation study on parameter \(c\). The results indicate that configurations with \(c=2\) or \(c=3\) at 50 steps achieve optimal performance in terms of both generation accuracy and human evaluation scores.}
    \vspace{-10px}
    \label{fig:9}
\end{figure}

Additionally, we perform an ablation study on the intensity parameter \(c\) of the AttnBoost mechanism, as illustrated in Fig.~\ref{fig:9}. The evaluation is conducted on the ScenePair dataset to measure text accuracy, supplemented by a human assessment phase that comprehensively evaluates both accuracy and aesthetic quality. The manual scoring uses a 100-point system, with 50 points allocated to text accuracy and 50 points to aesthetics, and the results are reported in percentage form. Tests are performed under different inference steps with varying intensity values. Results indicate that the highest scores are achieved when \(c = 2\) or \(c = 3\), while larger values of \(c\) do not lead to significant performance improvements. Although inference with 70 steps yields marginally better results, considering the trade-off between inference efficiency and computational cost, we select 50 steps with \(c=2\) and 50 steps as the default configuration, offering the most balanced solution in practice.

\begin{figure}
    \centering
    \includegraphics[width=1\linewidth]{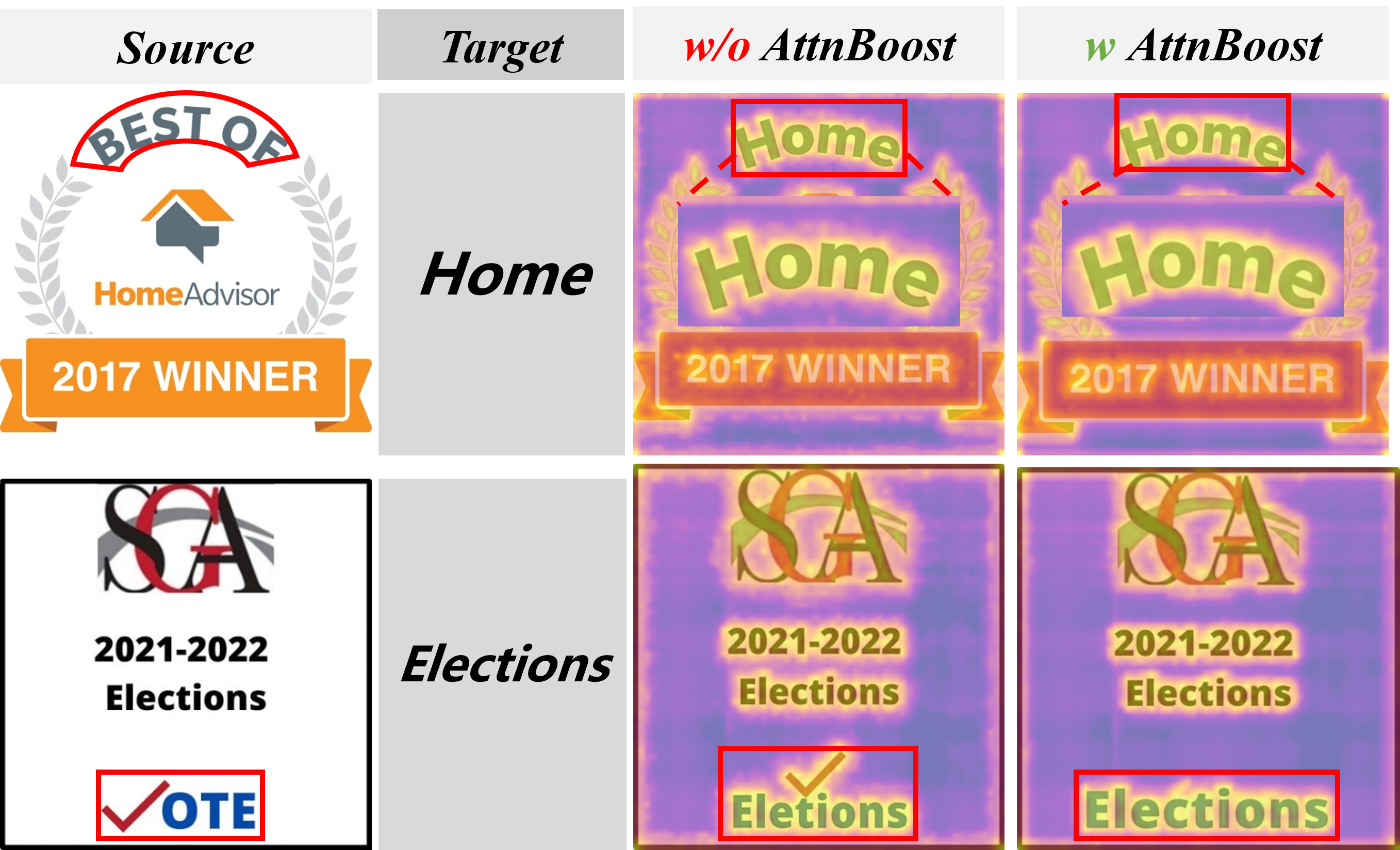}
    \caption{Visualized heatmaps of models with and without AttnBoost.}
    \label{fig:heat}
\end{figure}
To further validate the effectiveness of AttnBoost, we visualize its attention heatmaps in Fig.~\ref{fig:heat}. The results clearly demonstrate its role in enhancing textual accuracy.

\subsection{Ablation of Overshoot Scheduler}
\begin{table}
\centering
\footnotesize
\caption{Ablation of Overshoot scheduler on ScenePair.}
\vspace{-10px}
\scalebox{1}{\begin{tabular}{@{}llcc@{}}
\toprule
Method & Setting & ACC(\%) $\uparrow$ & NED $\uparrow$  \\ 
\midrule
TextFlow-Kontext & + Overs. + AttnBoost     & \underline{81.16} & \textbf{0.93}   \\
TextFlow-Kontext & + Overs.                 & 79.99 & 0.91  \\
TextFlow-Kontext & -                        & 78.72 & \underline{0.92}  \\
TextFLux         & + Overs.                 & \textbf{81.24} & 0.92  \\
TextFLux         & -                        & 80.40 & 0.91  \\
\bottomrule
\end{tabular}}
\label{tab:9}
\end{table}
Table~\ref{tab:9} ablates the overshoot scheduler on the ScenePair dataset. For TextFlow-kontext, introducing the overshoot scheduler alone boosts ACC from 78.72\% to 79.99\%, and further adding AttnBoost achieves the highest accuracy (81.16\% ACC and 0.93 NED). Similarly, textFlux also benefits from the overshoot scheduler, with ACC increasing from 80.40\% to 81.24\%. These results confirm that both the overshoot scheduler and AttnBoost contribute to improved textual accuracy.

\subsection{Time and GPU costs}

\begin{table}[t]
\centering
\footnotesize
\caption{Inference time and memory cost on A6000, accuracy on TamperScene.}
\vspace{-10px}
\scalebox{1}{\begin{tabular}{@{}llccc@{}}
\toprule
Model & Setting & Time(s) $\downarrow$ & Mme(GB) $\downarrow$ & ACC(\%) \\ 
\midrule
F-Kontext & Base & \textbf{260.21} & \textbf{40.12} & 17.79\\
F-Kontext & Base+TextFlow & 483.43 & 41.57 & \textbf{18.75}\\
\midrule
Longcat-Edit & Base & \textbf{136.54} & \textbf{37.24}  & 0.65\\
Longcat-Edit & Base+TextFlow & 252.62 & 38.67  & \textbf{20.95}\\
\midrule
FlowEdit & Base(FLUX.1dev) & 100.62 & 32.20 & 5.56\\
\bottomrule
\end{tabular}}
\label{tab:9}
\end{table}

Table~\ref{tab:9} reports inference time, GPU memory consumption, and accuracy on the TamperScene dataset using an A6000 GPU. For Flux-kontext, integrating TextFlow increases inference time from 260.21s to 483.43s and memory usage from 40.12GB to 41.57GB, while improving accuracy from 17.79\% to 18.75\%. For Longcat-edit, TextFlow raises time from 136.54s to 252.62s and memory from 37.24GB to 38.67GB, but delivers a dramatic accuracy boost from 0.65\% to 20.95\%. FlowEdit (based on FLUX.1-dev) serves as a baseline with 100.62s, 32.20GB, and 5.56\% accuracy. These results demonstrate that TextFlow achieves substantial gains in textual accuracy at the cost of moderate increases in computational resources, particularly for models that initially exhibit low text accuracy.

\subsection{Visualize Results and Limitations}
\begin{figure*}[h]
    \centering
    \includegraphics[width=1\linewidth]{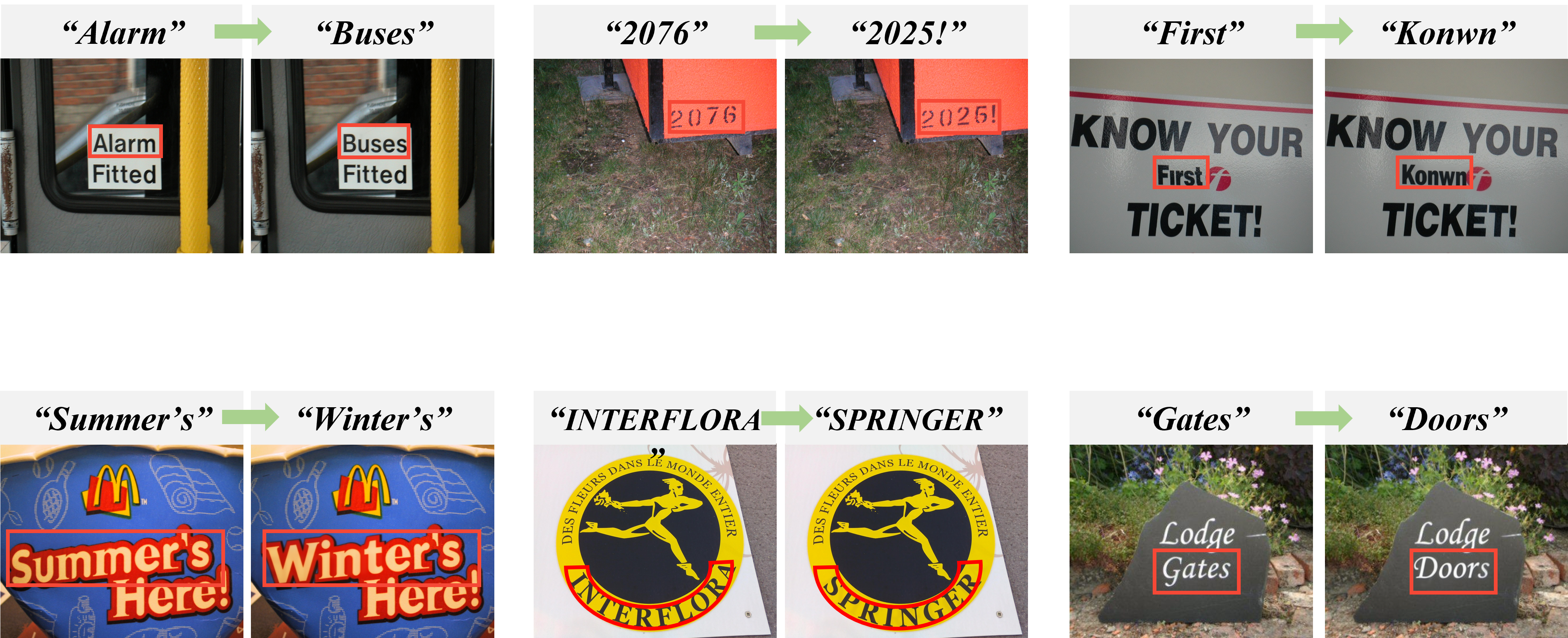}
    \caption{Additional visual results demonstrating robust performance across challenging scenarios, including artistic typography, small-font text, complex layouts, and special symbols.}
    \label{fig:11}
\end{figure*}

\noindent\textbf{Visualize Results}
Fig.~\ref{fig:11} demonstrates the editing performance of TextFlow across diverse challenging scenarios. Our method exhibits remarkable capability in handling special symbols, significant background luminance variations, fine-grained regions, artistic typography, and structurally complex layouts. Particularly noteworthy is its performance in the second row, where TextFlow successfully maintains style consistency and achieves accurate text rendering even when dealing with circular text arrangements, a particularly challenging case that requires sophisticated geometric adaptation.

\begin{figure*}
    \centering
    \includegraphics[width=1\linewidth]{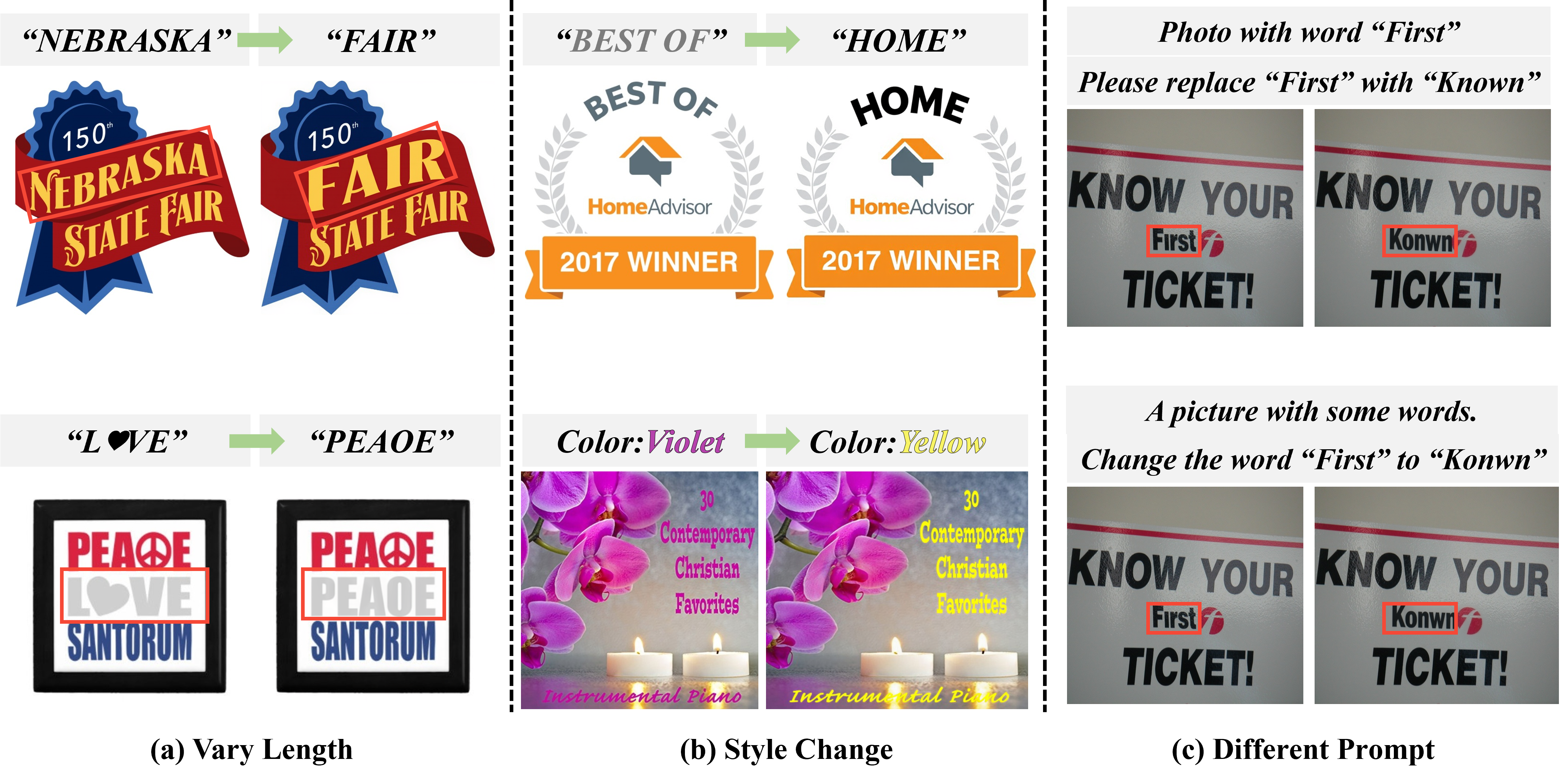}
    \caption{Extensive study on varying length, style change, and different prompts.}
    \label{fig:extensive}
\end{figure*}

Fig.~\ref{fig:extensive} (a) and (b) demonstrate the scalability of TextFlow across challenging editing tasks, including variations in word length and simple stylistic changes, highlighting its robust generalization to complex scenarios without task-specific tuning. Meanwhile, Fig.~\ref{fig:extensive} (c) illustrates its flexibility in responding to diverse user instructions, accurately performing edits according to different textual prompts, which underscores its adaptability for interactive applications.

\begin{figure*}[t]
    \centering
    \includegraphics[width=1\linewidth]{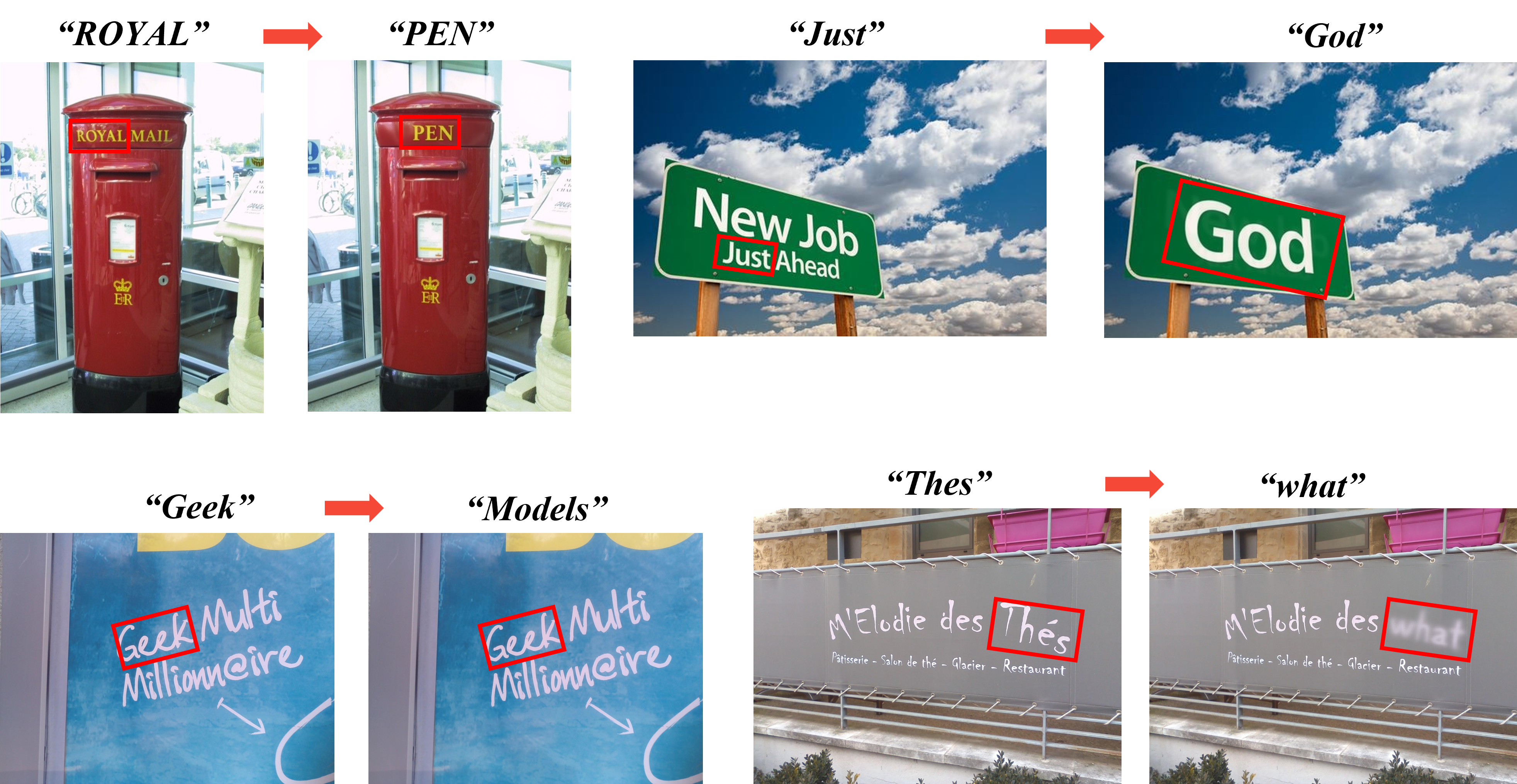}
    \caption{Limitations in editing accuracy and practical utility within complex scenarios.}
    \label{fig:12}
\end{figure*}

\noindent\textbf{Limitations}
Fig.~\ref{fig:12} illustrates certain limitations of our proposed method. The approach exhibits challenges in accurate word spatial localization and inter-word gap recognition, as evidenced by the case where “ROYAL” → “PEN” incorrectly merges both words into “PEN”. Additionally, the method demonstrates insufficient capability in handling images with perspective distortion, as shown in the “Just” → “God” example, where it erroneously modifies all textual elements while leaving residual background artifacts. For irregular character arrangements such as handwritten fonts, the editing process fails to achieve satisfactory results, as seen in the unsuccessful “Geek” → “Models” conversion. Furthermore, the method occasionally produces blurred rendering outputs, particularly evident in the “Thes” → “what” transformation, where character clarity is compromised.




\end{document}